%% file: main.tex
\documentclass[10pt,twocolumn,letterpaper]{article}

\usepackage{cvpr} 

\usepackage{lineno}
\usepackage{booktabs}
\usepackage{tabularx}
\usepackage{amssymb,amsmath}
\usepackage{mathrsfs}
\usepackage{forest}
\usepackage{xspace}
\usepackage{multirow}
\definecolor{cvprblue}{rgb}{0.21,0.49,0.74}
\usepackage[pagebackref,breaklinks,colorlinks,allcolors=cvprblue]{hyperref}

\def\paperID{xxxx} 
\def\confName{CVPR}
\def\confYear{2026}

\title{AnyTalker: Scaling Multi-Person Talking Video Generation \\with  Interactivity Refinement}

\author{
Zhizhou Zhong$^{1,2}$\quad
Yicheng Ji$^{2,3}$\quad
Zhe Kong$^{1}$\quad
Yiying Liu$^{2*}$\quad
Jiarui Wang$^{2}$\quad 
Jiasun Feng$^{2}$\quad \\
Lupeng Liu$^{2,4}$\quad
Xiangyi Wang$^{2,4}$\quad
Yanjia Li$^{2}$\quad
Yuqing She$^{2,4}$  
Ying Qin$^{4}$\quad 
Huan Li$^{3}$\quad \\
Shuiyang Mao$^{2}$\quad 
Wei Liu$^{2}$\quad
Wenhan Luo$^{1\dagger}$
\\
$^{1}$Hong Kong University of Science and Technology \quad
$^{2}$Video Rebirth \quad \\
$^{3}$Zhejiang University \quad
$^{4}$Beijing Jiaotong University
\\
{\tt\small Homepage: https://hkust-c4g.github.io/AnyTalker-homepage}
\\
{\tt\small Code: https://github.com/HKUST-C4G/AnyTalker}
}

\begin{document}

\twocolumn[{%
\renewcommand\twocolumn[1][]{#1}%
\maketitle
\includegraphics[width=\textwidth]{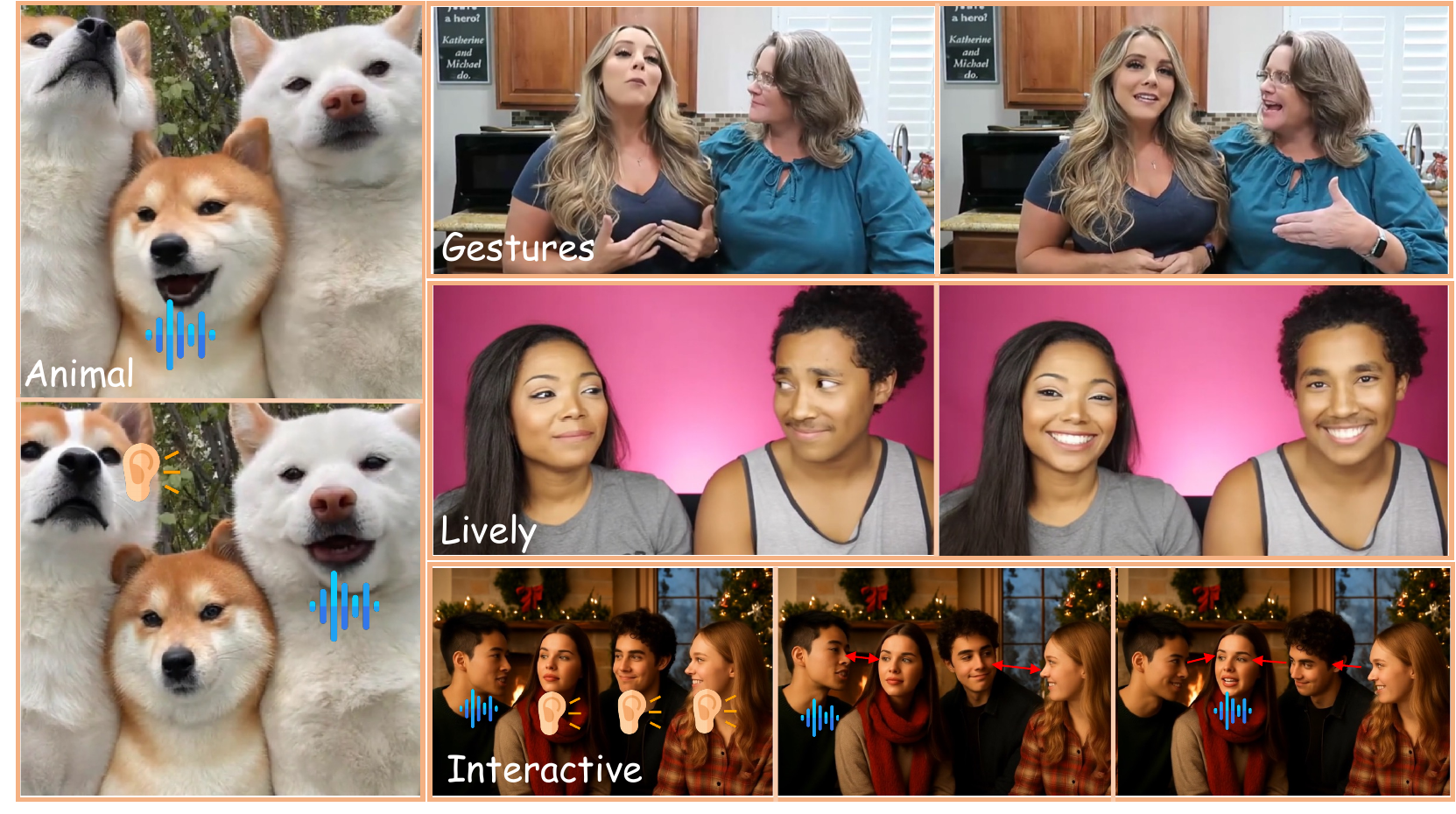}
\vspace{-1.5em}
\captionof{figure}{
We propose AnyTalker, a powerful audio-driven framework for interactive multi-person video generation. It can generate natural videos that are rich in gestures, lively emotions, and interactivity, and can freely generalize to arbitrary IDs or even non-human cases.
\vspace{1em}}
\label{fig:teaser}
}]

\def\thefootnote{}\footnotetext{$^*$ Project leader.}
\def\thefootnote{}\footnotetext{$^\dagger$ Corresponding author.}

\input{sec/0_abstract}    
\input{sec/1_intro}
\input{sec/2_related_work}

\input{sec/3_method}

\input{sec/4_evaluation}
\input{sec/5_exp}

\input{sec/6_conclusion}
{
    \small
    \bibliographystyle{ieeenat_fullname}
    \bibliography{main}
}
\input{sec/X_suppl}


\end{document}

%% file: sec/0_abstract.tex
\begin{abstract}
Recently, multi-person video generation has started to gain prominence. 
While a few preliminary works have explored audio-driven multi-person talking video generation, they often face challenges due to the high costs of diverse multi-person data collection and the difficulty of driving multiple identities with coherent interactivity.
To address these challenges, we propose AnyTalker, a multi-person generation framework that features an extensible multi-stream processing architecture.
Specifically, we extend Diffusion Transformer's attention block with a novel identity-aware attention mechanism that iteratively processes identity–audio pairs, allowing arbitrary scaling of drivable identities.
Besides, training multi-person generative models demands massive multi-person data.
Our proposed training pipeline depends solely on single-person videos to learn multi-person speaking patterns and refines interactivity with only a few real multi-person clips.
Furthermore, we contribute a targeted metric and dataset designed to evaluate the naturalness and interactivity of the generated multi-person videos.
Extensive experiments demonstrate that AnyTalker achieves remarkable lip synchronization, visual quality, and natural interactivity, striking a favorable balance between data costs and identity scalability.
\end{abstract}

%% file: sec/1_intro.tex
\section{Introduction}
\label{sec:intro}
In the era of digital media, video creation has emerged as a crucial component of media platforms. 
Podcasts, live-streaming sales, and entertainment programs often feature rich multi-person interactions, resulting in a growing demand for multi-person video generation.
Despite the advent of large-scale video generation models~\cite{blattmann2023stable, yangcogvideox, kong2024hunyuanvideo, wan2025wan}, which have furnished robust backbone architectures for audio-driven talking video generation methods~\cite{lin2025omnihuman, gan2025omniavatar, jiang2025omnihuman, yang2025infinitetalk, cui2025hallo3, chen2025hunyuanvideo, ji2025sonic, gao2025wan, kong2025let}, enabling the creation of realistic lip movements for single subjects, these models still struggle to accommodate the complexities of multi-person interactions.

This limitation has prompted multi-person driving approaches that scale to arbitrary numbers of persons, handle multi-stream signals, and enable differentiated control over each subject.
Despite recent advances~\cite{wang2025interacthuman, huang2025bind, kong2025let} that propose solutions for handling multi-stream audio signals, these methods typically require hundreds to thousands of hours of meticulously curated multi-person data, leading to prohibitive collection costs and limited reproducibility. 
Specifically, the challenge of gathering training data for multi-person scenarios is heightened by their complex dynamics, including turn-taking, role-switching, and nonverbal cues like eye gaze, which complicate data annotation. 
Most existing audio-visual datasets~\cite{zhang2021flow, xie2022vfhq, cui2025hallo3, meng2025echomimicv2, li2025openhumanvid, chung2018voxceleb2} focus on single-person monologues or isolated facial animations, thereby limiting their applicability for training the audio-driven multi-person generation model. 
In addition, current multi-person driving approaches frequently struggle to model genuine interactivity, leading to unnatural results.

To address the aforementioned challenges, we introduce an innovative driving framework named AnyTalker. 
Built on the pre-trained video diffusion model~\cite{wan2025wan}, AnyTalker achieves impressive multi-person video driving with remarkably low data costs and a distinct emphasis on interactivity, as illustrated in \cref{fig:teaser}.
The central idea is to leverage low-cost single-person data for scalable multi-person video driving and refine interactivity with just a small amount of multi-person data (as little as 12 hours). 
Specifically, AnyTalker supports extending the number of drivable identities (IDs) to arbitrary numbers, with guaranteed interactivity among all IDs. 
To accommodate multi-stream control signals, we design an extensible audio-to-face in-context attention mechanism supporting any number of IDs and audio inputs.
The training process is divided into two stages based on the types of data used, with the number of speaker IDs in individual data samples evolving from one to many.
In the first stage, we randomly concatenate single-person talking videos along the horizontal dimension to simulate multi-person talking scenarios, ensuring that the model acquires a baseline capacity for multi-person speaking patterns. 
In the second stage, we fine-tune the model using a small amount of multi-person data to enhance its interactive capabilities.

Commonly used single-person talking head benchmarks~\cite{zhang2021flow, xie2022vfhq, zhu2022celebv} lack multi-person interactions, rendering them unsuitable for assessing multi-person generation methods.
While InterActHuman~\cite{wang2025interacthuman} offered a related benchmark, it focuses on single speakers, limiting its utility for interaction analysis. 
To address this limitation, we introduce a meticulously annotated benchmark featuring videos of two individuals engaging in both speech and eye contact, with fine-grained labels that mark the speaking and listening intervals, thus facilitating the assessment of interactions during listening states.
Additionally, we firstly introduce a novel metric to evaluate interactivity by measuring the activity of eye keypoints during listening periods. 
The proposed benchmark and metric will fill the gap in the assessment of interactivity in multi-person generation methods, benefiting future research in this area.

The main contributions are summarized as follows:
(1) We present an extensible multi-stream processing architecture for multi-person generation that can scale drivable identities arbitrarily.
(2) A novel two-stage training pipeline is introduced for the model to learn multi-speaker speaking patterns from single-person data and achieve seamless inter-identity interactions via multi-person data refinement.
(3) We propose a new metric that quantitatively evaluates multi-person interactivity for the first time, accompanied by a tailored benchmark dataset for thorough assessment.
(4) Comprehensive experiments demonstrate that AnyTalker achieves state-of-the-art performance and strikes a favorable balance among identity scalability, interactivity, lip synchronization, and data cost.

%% file: sec/2_related_work.tex
\section{Related Work}
\label{sec:related-work}

\begin{figure*}[t]
    \centering
    \includegraphics[width=\linewidth]{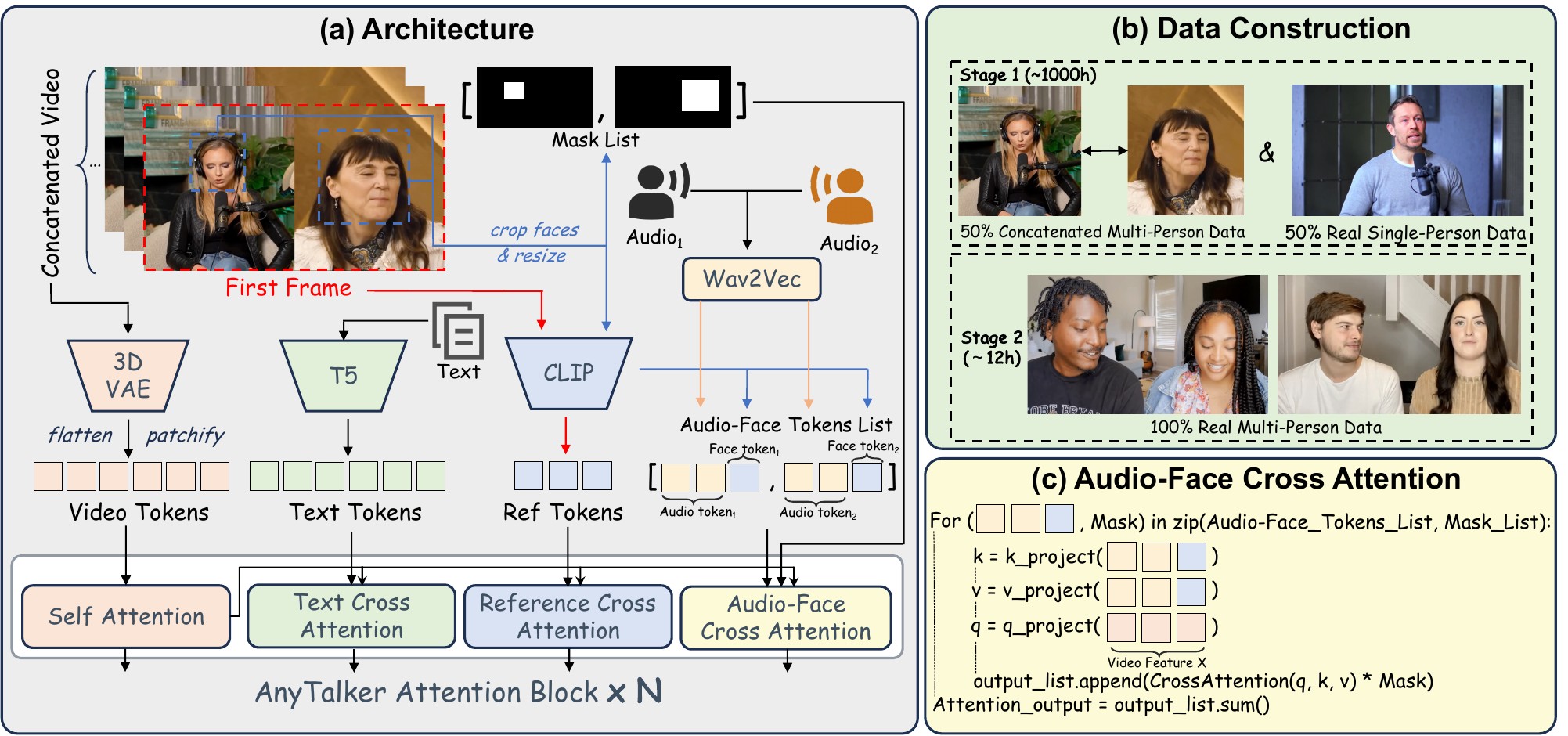}
    \caption{
    (a) The architecture of AnyTalker, which incorporates a novel multi-stream audio processing layer, Audio-Face Cross Attention, enables the handling of multiple facial and audio inputs. 
    (b) The training of AnyTalker is divided into two stages: the first stage uses concatenated multi-person data derived from single-person data mixed with single-person data to learn accurate lip movements; the second stage employs authentic multi-person data to enhance the interactivity in generated videos. 
    (c) The detailed implementation of Audio-Face Cross Attention, a recursively callable structure that applies masking to the output using face masks.
    }
    \label{fig:method}
\end{figure*}

\subsection{Audio-driven Talking Video Generation}
Recent key advancements~\cite{ho2020denoising, rombach2022high, ho2021classifier, zhang2023adding, peebles2023scalable} in text-to-image fields have significantly catalyzed progress in downstream applications. 
Early efforts like EMO~\cite{tian2024emo} extend pre-trained diffusion text-to-image models~\cite{rombach2022high} to end-to-end audio-driven virtual-human video generation~\cite{jiangloopy, chen2025echomimic, xu2024hallo, yee2025synchrorama, wang2024v, nazarieh2024portraittalk, lin2025cyberhost, chen2025midas}. 
They typically integrate modules for temporal attention~\cite{guoanimatediff}, identity control via ReferenceNet~\cite{zhu2023tryondiffusion, hu2024animate, kong2025profashion, zhang2025learning}, and audio-conditioning using pre-trained audio processing models~\cite{radford2023robust, baevski2020wav2vec, schneider2019wav2vec}, with conditional signals injected via cascaded attention layers~\cite{vaswani2017attention}. 
Video foundation models~\cite{blattmann2023stable, yangcogvideox, kong2024hunyuanvideo, wan2025wan, Zhang_2025_CVPR} enable end-to-end, audio-driven virtual-human models~\cite{meng2025echomimicv3, cui2025hallo3, ji2025sonic, tu2025stableavatar, lin2025omnihuman, yang2025infinitetalk, gao2025wan, jiang2025omnihuman}. 
These works largely employ audio-injection practices from earlier methods~\cite{tian2024emo, jiangloopy} and report breakthroughs in long-video generation~\cite{yang2025infinitetalk, cui2025hallo3}, full-body synthesis~\cite{lin2025omnihuman}, 
driving non-human cases~\cite{gao2025wan, gan2025omniavatar}, synchronized lip movements~\cite{ji2025sonic}, and coherent hand movements~\cite{meng2025echomimicv3}.
Consequently, they surpass earlier GAN-based generation methods~\cite{zhang2023sadtalker, xu2024vasa, guo2024liveportrait, zhang2024musetalk} and image diffusion-based techniques~\cite{rombach2022high, peebles2023scalable} in terms of both quality and controllability. 
However, most methods still target single-person scenarios; in multi-person scenarios, they tend to synchronize identical motions or lip movements across all speakers, with restricted multi-person interaction.

\subsection{Multi-person Video Generation} Multi-person video generation has advanced rapidly through specialized architectures, including portrait video generation~\cite{wang2025fantasyportrait, meng2025identity}, dancing video generation~\cite{xuetowards}, and talking video generation~\cite{huang2025bind, kong2025let, wang2025interacthuman}. 
Within the audio-driven talking-head axis, Bind-your-Avatar~\cite{huang2025bind} introduces a fine-grained Embedding Router that binds ``who” with ``what they speak”. InterActHuman~\cite{wang2025interacthuman} trains a mask predictor to identify which body regions to activate, enabling targeted control. 
MultiTalk~\cite{kong2025let} proposes Label Rotary Position Embedding~\cite{su2024roformer} to address audio–person binding. 
All three approaches above rely on expensive data, ranging from hundreds to thousands of hours of multi-person talking data. 
Some models~\cite{chen2025hunyuanvideo, ma2025playmate2} trained on single-person data generalize to multi-person driving through specialized controllers: HunyuanVideo-Avatar~\cite{chen2025hunyuanvideo} leverages a Face-Aware Audio Adapter to activate attention across different characters selectively, and Playmate2~\cite{ma2025playmate2} uses token-level masking within a classifier-free guidance framework to realize similar binding. 
Despite enabling multi-person outputs, these approaches might yield fragmented interactions across distinct characters, with limited interactivity between individuals.

In this work, AnyTalker explores the potential of learning multi-person speaking patterns from single-person data and designs an extensible multi-stream audio processing attention architecture, achieving a favorable balance among interactivity, identity scalability, and lip synchronization in the generated videos with low training data cost. 

%% file: sec/3_method.tex
\section{Method}
\label{sec:method}
The overall framework of the proposed AnyTalker is illustrated in~\cref{fig:method}. 
AnyTalker inherits certain architectural components from the Wan I2V model~\cite{wan2025wan}.
To handle multi-stream audio and identity input, we introduce a specialized multi-stream processing structure, termed as the \textit{Audio-Face Cross Attention}~(AFCA), which will be further described in~\cref{subsec:audio-face-attn}. 
Our training pipeline is divided into two stages, which are summarized in~\cref{subsec:training-strategy}.

\begin{figure}[t]
    \centering
    \includegraphics[width=0.9\linewidth]{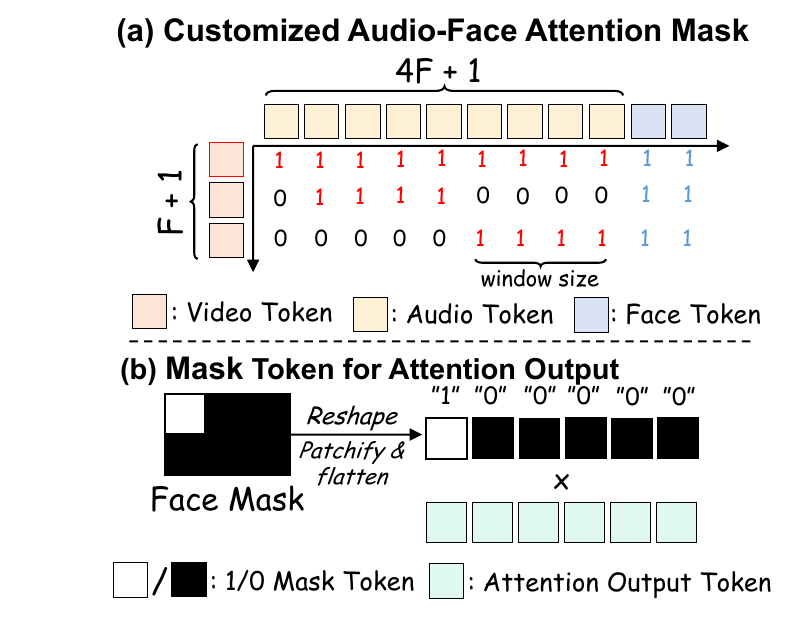}
    \caption{(a) Mapping of video tokens to audio tokens, facilitated by a custom attention mask. 
    Every 4 audio tokens are bound to 1 video token, except for the first. 
    (b) Mask token used for output masking in Audio-Face Cross Attention.}
    \label{fig:token}
\end{figure}

\subsection{Preliminaries}
As a DiT-based model, AnyTalker tokenizes the 3D VAE features $f_{video}$ through patchifying and flattening, whereas the text features $f_{text}$ are generated by the T5 encoder~\cite{raffel2020exploring}. 
Additionally, AnyTalker incorporates Reference Attention Layer, a cross-attention mechanism that leverages the CLIP image encoder~\cite{radford2021learning} \(E_{\text{CLIP}} \) to extract features $f_{ref}$ from the first frame of the video. Wav2Vec2~\cite{baevski2020wav2vec} is also applied to extract the audio feature $f_{audio}$. 
The overall input features $f_{input}$ can be written as
\begin{equation}
    f_{input} = [f_{video}, f_{text}, f_{ref}, f_{audio}].
\end{equation}
\label{eq:feature}
Consistent with the Wan model, all attention layers are connected to the final output FFN layer (omitted in~\cref{fig:method}).

\subsection{Audio-Face Cross Attention}
\label{subsec:audio-face-attn}
To enable multi-person talking, the model must be capable of handling multi-stream audio inputs.
Potential solutions may include the L-RoPE technique used in MultiTalk~\cite{kong2025let}, which assigns unique labels and biases to different audio features.
However, the range of these labels needs to be explicitly defined, which limits its scalability.
Considering this, we design a more extensible structure to drive multiple IDs and enable accurate control in a scalable manner.
As depicted in~\cref{fig:method}~(a) and (c), we introduce a specialized structure named Audio-Face Cross Attention (AFCA). 
The structure can iterate through a loop multiple times, contingent upon the number of input face-audio pairs. 
As illustrated in~\cref{eq:attn} and~\cref{fig:method} (c), it enables flexible processing of diverse audio and identity inputs, with summed outputs from each iteration yielding the final attention output.

\noindent \textbf{Audio Token Modeling.}
We employ Wav2Vec2~\cite{baevski2020wav2vec} to encode audio features. 
The first latent frame attends to all audio tokens, whereas each subsequent latent frame focuses only on a local temporal window corresponding to four audio tokens. 
This structured alignment between video and audio streams is achieved by applying a Temporal Attention Mask \( M_{\text{temporal}} \), as explicitly shown in~\cref{fig:token}~(a). 
Furthermore, to enable comprehensive information integration, each audio token \( f_{\text{audio}} \) utilized in the AFCA computation is concatenated with a face token \( f_{\text{face}} \) encoded by \( E_{\text{CLIP}} \).
This concatenation allows all video query tokens \( Q_{\text{video}} \) to attend to different pairs of audio and face information effectively, as computed below:
\begin{equation}
\label{eq:af-attention}
\begin{aligned}
    K_{\text{af}} &= \text{Concat}(f_{\text{audio}}, f_{\text{face}})\cdot W_K, \\
    V_{\text{af}} &= \text{Concat}(f_{\text{audio}}, f_{\text{face}})\cdot W_V, \\
    \text{Attn}_{\text{out}} &= \text{MHCA}(Q_{\text{video}}, K_{\text{af}}, V_{\text{af}}, M_{\text{temporal}}).
\end{aligned}
\end{equation}
Here, MHCA denotes \textit{Multi-Head Cross Attention}, while $W_K$ and $W_V$ represent the key matrix and value matrix, respectively.
The attention output $\text{Attn}_{\text{out}}$ will later be refined by the face mask token, as described in~\cref{eq:mask}.

\noindent \textbf{Face Token Modeling.}
The facial image is obtained by online cropping the first frame of the selected video clip using InsightFace~\cite{deng2020retinaface} during training, while the facial mask $M_{\text{face}}$ is precomputed offline to cover the maximum extent of the face mask across the entire video, i.e., the \textit{global} face bounding box. 
This mask ensures that facial movements will never exceed this region, preventing the mask from incorrectly activating video tokens after the reshape and flatten operations shown in~\cref{fig:token}~(b), especially for videos with significant facial displacements.
This mask, which shares the same dimensions as \( \text{Attn}_{\text{out}} \), can be directly employed for element-wise multiplication to compute the \textit{Audio-Face Cross Attention} output, as formulated below:
\begin{equation}
\begin{aligned}
{M}_\text{token} &= \text{Patchify}( \text{Flatten}(M_\text{face})), \\
\text{AFCA}_\text{out} &= {M}_\text{token} \odot \text{Attn}_{\text{out}}.  \label{eq:mask}
\end{aligned}
\end{equation}
Consequently, the hidden state $H_{i}$ of each I2V DiT block can be formulated as
\begin{equation}
\begin{aligned}
H_{i}^{'} &= H_{i} + \text{AFCA}_\text{out}^{(1)}
+ \cdots + \text{AFCA}_\text{out}^{(n)}, \label{eq:attn}
\end{aligned} 
\end{equation}
where $i$ represents the layer index of the attention block, and $n$ denotes the number of IDs.
Note that all $\mathrm{AFCA}_{\mathrm{out}}$ terms are produced by the \emph{same} AFCA layer with shared parameters. 
The AFCA computation is applied $n$ times iteratively, once for each individual.
This architecture enables the number of drivable IDs to scale arbitrarily.

\begin{figure*}[t]
    \centering
    \includegraphics[width=\linewidth]{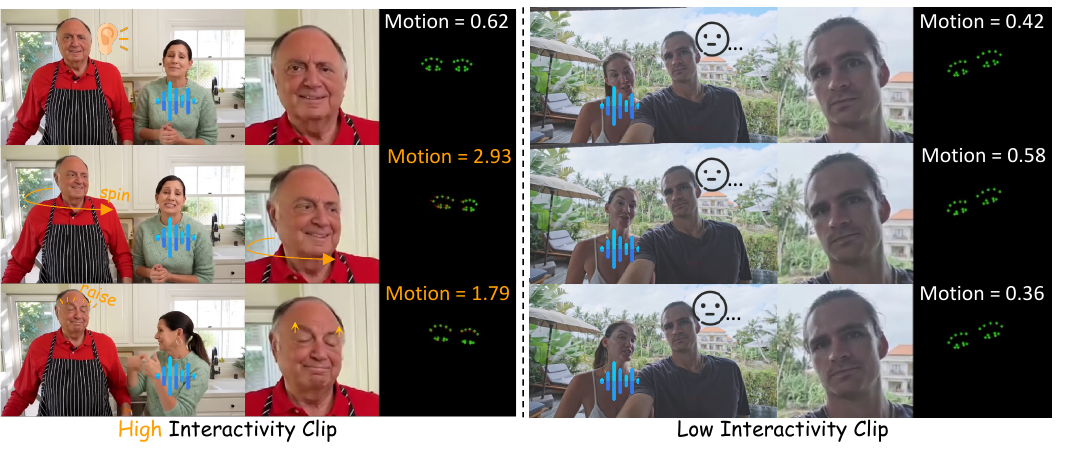}
    \caption{
    Two video clips from InteractiveEyes with ${Motion}$ score (px): left shows original video, right shows cropped face and eye landmarks. Head turn toward the speaker or eyebrow raise will increase $Motion$ and Interactivity; sustained stillness keeps both low.
    }
    \label{fig:interactivity}
\end{figure*}

\subsection{Training Strategy}
\label{subsec:training-strategy}
AnyTalker explores the potential of single-person data for learning multi-person speaking patterns, with low-cost single-person data comprising the majority of training data.

\noindent \textbf{Single-Person Data Pretraining.}
We train the model using both standard single-person data and synthetic two-person data generated by horizontal concatenation. 
With an equal 50\% probability, each batch of data is randomly configured to either two-person or single-person mode, as depicted in~\cref{fig:method}~(b).
In two-person mode, each sample within the batch is horizontally concatenated with the data at the next index, along with its corresponding audio. 
This approach keeps the batch size identical across the two modes for every data batch.
Additionally, we have predefined several generic text prompts that describe dual people speaking when data concatenation happens.

Although the above data construction strategy enhances the model's ability to localize audio-visual features within local regions and learn speaking patterns of dual speakers, completely omitting the single-person data is not feasible. 
Doing so would significantly degrade the model's performance on generating accurate lip movements, leading to unstable driving results, which we later discuss in~\cref{tab:ab-data}.


%
\noindent \textbf{Multi-person Data Refinement.}
In the next stage, we refine the model using a small amount of authentic multi-person data to enhance the interactivity within different IDs.
Although our training data contains only interactions between two identities, we surprisingly find that our model equipped with the AFCA module naturally generalizes to scenarios with more than two IDs, as shown in~\cref{fig:teaser}.
We speculate that this is because the AFCA mechanism enables learning general patterns of human interaction, including not only accurate lip-syncing to the audio but also listening and responsive behaviors to other IDs' speaking actions.

To construct high-quality multi-person training data, we construct a rigorous quality control pipeline, using InsightFace~\cite{deng2019arcface} to ensure two faces in most frames, audio diarization~\cite{Plaquet23} to separate audio and ensure there is only one or two speakers, optical flow~\cite{karaev2024cotracker} to filter excessive motion, and Sync scores~\cite{chung2016out} to pair audio with faces. 
More details about this pipeline are in the supplementary material.
This pipeline yields a total of $12$ hours of high-quality dual-person data, which is a small amount compared to previous methods~\cite{kong2025let, wang2025interacthuman, ma2025playmate2}. 
As the design of AnyTalker's AFCA Layer inherently supports multi-ID inputs, two-person data is fed into the model in the same format as the concatenated data in the first stage, with no extra processing required.

To summarize, the single-person data training process enhances the model's lip-syncing capability and generation quality, while also learning a generalized multi-person speaking pattern. Subsequently, lightweight multi-person data refinement compensates for the real interactions that cannot be fully covered by single-person data.

%% file: sec/4_evaluation.tex
\section{Interactivity Evaluation}
\label{sec:evaluation}
Despite progress, the prevailing evaluation benchmarks~\cite{zhang2021flow, xie2022vfhq, zhu2022celebv} for single-person talking head generation are inadequate for assessing natural interactions among characters.
Although InterActHuman~\cite{wang2025interacthuman} introduces a comparable benchmark, its test set is limited to scenarios with only one speaker, which is not conducive to evaluating interactions among multiple characters.
To fill this gap, we have curated a collection of videos featuring two distinct identities, sourced from the web for evaluation purposes.

\subsection{Dataset Construction}
\begin{figure}[t]
    \centering
\includegraphics[width=0.7\linewidth]{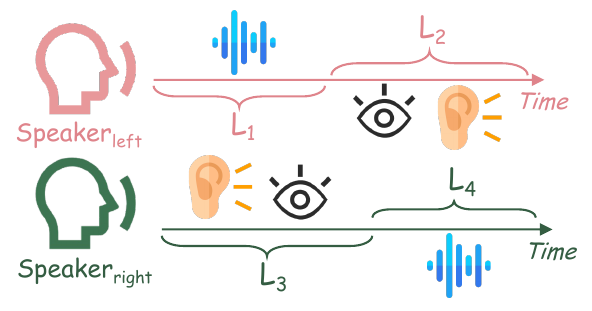}
    \caption{
    Listening and speaking periods of each speaker.
    }
    \label{fig:time}
\end{figure}

We select interactive two-person videos to construct the video dataset, named \textit{InteractiveEyes}.
Two clips of these videos are illustrated in~\cref{fig:interactivity}. 
Each video is approximately 10 seconds in duration and showcases exactly two faces throughout the entire segment. 
Furthermore, through a meticulous manual process, we segment the audio of each video to ensure that the majority of the videos capture scenes of both individuals engaging in \textit{speaking} and \textit{listening}, as well as a variety of rich eye interaction scenarios, as shown in~\cref{fig:time}. 
We have also ensured that each video includes instances of mutual gaze and head movements to provide authentic references.

\subsection{Proposed Interactivity Metric}
In addition to this dataset, we introduce a novel metric, the eye-focused \textit{Interactivity}, designed to assess the natural interaction between speakers and listeners. 
Since eye interaction is a fundamental and spontaneous behavior in conversational contexts, we use it as a key indicator of interactivity.
Drawing inspiration from the Hand Keypoint Variance (HKV) metric employed in CyberHost~\cite{lin2025cyberhost}, we propose a quantitative evaluation of the interaction by tracking the motion amplitude of eye keypoints.

To achieve this, we define $Motion$ on the sequence of face-aligned eye keypoints extracted from the generated frames, where $S$ denotes the frame sequence and $E$ the eye keypoints.
The $Motion$ is calculated as follows
\begin{equation}
    Motion = \frac{1}{|S| - 1} \sum_{j=1}^{|S| - 1} \left( \frac{1}{|E|} \sum_{i=1}^{|E|} \vert E_{i,j+1} - E_{i,j}\vert \right).
\end{equation}
Here, $i$ and $j$ denote the eye keypoint index and the frame index, while \( E_{i,j} \) denotes eye keypoints present in each frame.
This formula intuitively computes the displacement and rotation of the eye region.
We then calculate the motion during listening periods. 
The reason is, most generation methods perform well when activating the speaking subject, but the listening subject often appears rigid. 
Therefore, evaluating during listening periods is more targeted and valuable.
The lengths of the listening and speaking periods of each person are described in~\cref{fig:time}, denoted as \( L_1, L_2, L_3, L_4 \), respectively.
To quantify the responsiveness of the generated avatars, we compute the average motion intensity during the listening phases $L_2$ and $L_3$:
\begin{equation}
\begin{aligned}
\text{Interactivity} &= \frac{L_2 \cdot Motion_{L2} + L_3 \cdot Motion_{L3}}{L_2 + L_3}.
\end{aligned}
\end{equation}
This metric measures the interactivity effectively in the generated multi-character videos. 
As~\cref{fig:interactivity} suggests, the proposed metric aligns well with human perception: static or sluggish eye movements receive low $Motion$ scores, while head turns and eyebrow raises increase the score thus indicating higher interactivity.
Moreover, to avoid misjudging abnormal eye movements, we implemented an exclusion algorithm detailed in the supplementary materials.

\begin{table*}[t]
    \centering
    \small
    \caption{Quantitative comparison with other competing methods on HDTF~\cite{zhang2021flow} and VFHQ~\cite{xie2022vfhq} benchmark.
    Here, OmniHuman-1.5$^{*}$~\cite{jiang2025omnihuman} refers to its ``Master Mode'' version accessed via the JiMeng platform~\cite{jimeng_platform}, which currently does not support multi-person generation.
    }
    \begin{tabular}{ccccccccccc}
        \toprule
        \multirow{2}{*}{\textbf{Method}} & \multicolumn{2}{c}{\textbf{Supported Generation Scope}} & \multicolumn{4}{c}{\textbf{HDTF}} & \multicolumn{4}{c}{\textbf{VFHQ}} \\
        \cmidrule(lr){2-3} \cmidrule(lr){4-7} \cmidrule(lr){8-11}
        & \textbf{multi-person} & \textbf{body} & \textbf{Sync-C$\uparrow$} & \textbf{FID$\downarrow$} & \textbf{FVD$\downarrow$} & \textbf{ID$\uparrow$} & \textbf{Sync-C$\uparrow$} & \textbf{FID$\downarrow$} & \textbf{FVD$\downarrow$} & \textbf{ID$\uparrow$} \\
        \midrule
        AniPortrait~\cite{wei2024aniportrait} &  $\times$ & $\times$ & 3.44 & 18.74 & 241.84 & \underline{0.94} & 2.63 & 28.54 & 269.24 & \textbf{0.95} \\
        FantasyTalking~\cite{wang2025fantasytalking} &  $\times$ & $\checkmark$ & 3.97 & 14.93 & 166.79 & 0.93 & 3.57 & 24.83 & 272.13 & \underline{0.94} \\
        StableAvatar~\cite{tu2025stableavatar} &  $\times$ & $\checkmark$  & 4.11 & 14.67 & 166.44 & 0.91 & 3.53 & 22.91 & 275.73 & 0.88 \\
        EchoMimic~\cite{chen2025echomimic} &  $\times$ & $\times$ & 5.23 & 61.53 & 381.55 & \textbf{0.95} & 4.87 & 58.72 & 486.75 & 0.86 \\
        Hallo3~\cite{cui2025hallo3} & $\times$ & $\times$ & 7.53 & 17.12 & 195.61 & 0.91 & 6.32 & 41.26 & 371.24 & 0.91 \\
        Sonic~\cite{ji2025sonic} &  $\times$ & $\times$ & 7.81 & 52.96 & 286.12 & \textbf{0.95} & 7.71 & 36.68 & 385.37 & 0.89 \\
        OmniHuman-1.5$^{*}$~\cite{jiang2025omnihuman} &  $\times$ & $\checkmark$   & 7.23  & 35.26  & 173.23  & 0.90  & 7.67  & 35.36  & 283.39  & 0.90  \\
        MuitiTalk~\cite{kong2025let} &  $\checkmark$ & $\checkmark$ & \underline{8.91} & \textbf{13.54} & \underline{162.58} & 0.93 &  \underline{7.77} & 24.25 & \textbf{243.66} & \underline{0.94} \\
        \midrule
        AnyTalker-1.3B & $\checkmark$ & $\checkmark$ & 6.85 & 14.47 & 218.01 & 0.91 & 5.81 & \underline{21.88} & \underline{267.08} & 0.91 \\
        AnyTalker-14B & $\checkmark$ & $\checkmark$ & \textbf{9.05} & \underline{13.84} & \textbf{160.87} & \underline{0.94} & \textbf{7.79} & \textbf{20.99} & 290.73 & \underline{0.94} \\
        \bottomrule
    \end{tabular}
    \label{tab:single-benchmark}
\end{table*}

\begin{table}[t]
    \centering
    \small
    \caption{
        Quantitative comparison with other competing methods on the multi-person benchmark, InteractiveEyes.
    }
    \begin{tabular}{ccccc}
    \toprule
    \textbf{Method} & \textbf{Interactivity$\uparrow$} & \textbf{Sync-C$^{*}$$\uparrow$} & \textbf{FVD$\downarrow$} \\
    \midrule
    Ground Truth & 0.77 & 6.01 & 0 \\
     \midrule
    Bind-Your-Avatar~\cite{huang2025bind} & 0.45 & 3.03 & 695.58  \\
    MuitiTalk~\cite{kong2025let} & 0.49 & \underline{6.88} & 500.03  \\
    AnyTalker-1.3B & \underline{0.97} & 4.56 & \underline{467.84}  \\
    AnyTalker-14B & \textbf{1.01} & \textbf{6.99} & \textbf{424.15}   \\
    \bottomrule
    \end{tabular}
    \label{tab:multi-benchmark}
\end{table}

%% file: sec/5_exp.tex
\section{Experiments}
\label{sec:exp}

\noindent \textbf{Dataset.} 
We expand single-person datasets~\cite{zhang2021flow, xie2022vfhq, cui2025hallo3, zhu2022celebv, chung2018voxceleb2} with internet-collected data, yielding roughly 1,000 hours for first-stage training, and also gather two-person conversation clips for the second-stage training, retaining only about 12 hours after filtering.
Evaluations are conducted on two types of benchmarks: (i) standard talking-head benchmarks HDTF~\cite{zhang2021flow} and VFHQ~\cite{xie2022vfhq}, and (ii) our self-collected multi-person conversation dataset (head-and-body, both identities speak).  
We then select 20 videos from each benchmark, rigorously ensuring that their identities do not appear in the training set.

\label{subsec:metric}

\noindent \textbf{Implement Details.}
To comprehensively evaluate our method, we train two models of different sizes: Wan2.1-1.3B-Inp and Wan2.1-I2V-14B~\cite{wan2025wan}, which serve as the foundational video diffusion models for our experiments. 
In all stages, the text~\cite{raffel2020exploring}, audio~\cite{baevski2020wav2vec}, and image~\cite{radford2021learning} encoders, as well as the 3D VAE, remain frozen.
The DiT main network, including the newly added AFCA layers, has all its parameters open for training. 
Stage 1 pretrains at $2 \times 10^{-5}$ learning rate; stage 2 fine-tunes at $5 \times 10^{-6}$.
All models are optimized with AdamW~\cite{loshchilov2017decoupled} on 32 NVIDIA H200 GPUs.

\noindent \textbf{Evaluation Metrics.}
For the single-person benchmark, we employ several commonly used metrics: the Fréchet Inception Distance (FID)~\cite{heusel2017gans} and the Fréchet Video Distance (FVD)~\cite{unterthiner2019fvd} to assess the quality of the generated data, Sync-C~\cite{chung2016out} to measure the synchronization between audio and lip movements, and ID similarity~\cite{deng2019arcface} calculated between the first frame and the remaining frames.

For the multi-person benchmark, we evaluate from different dimensions. 
The newly introduced metric, termed \textit{Interactivity}, serves as the primary metric for assessment. 
For the FVD metric, the calculation is similar to that in the single-person benchmark. 
For the Sync-C metric, we refine its calculation as Sync-C$^{*}$ to focus only on the lip synchronization during each character's speaking periods, thereby avoiding the influence of long listening segments on the final lip synchronization score, specifically,
\begin{equation}
\text{Sync-C}^{*} = \frac{L_1 \cdot \text{Sync-C}_{L1} + L_4 \cdot \text{Sync-C}_{L4}}{L_1 + L_4}.
\end{equation}
Here, $L_1$ and $L_4$ denotes the speaking phases depicted in~\cref{fig:time}.

\noindent \textbf{Comparsion Methods.}
We compare AnyTalker with several state-of-the-art talking video generation methods.
For single-person generation, we compare with AniPortrait~\cite{wei2024aniportrait}, EchoMimic~\cite{chen2025echomimic}, Hallo3~\cite{cui2025hallo3}, Sonic~\cite{ji2025sonic}, FantasyTalking~\cite{wang2025fantasytalking}, StableAvatar~\cite{tu2025stableavatar}, OmniHuman-1.5~\cite{jiang2025omnihuman}, and MultiTalk~\cite{kong2025let}.
For multi-person generation, we choose Bind-Your-Avatar~\cite{huang2025bind} and MultiTalk~\cite{kong2025let} for quantitative and qualitative comparison.

\subsection{Comparison with SOTA methods}
\noindent \textbf{Quantitative Comparison.}
To begin with, we compare AnyTalker with several single-person generation methods to verify its excellent single-person driving capability. 
The quantitative results are shown in~\cref{tab:single-benchmark}. 
Despite not being specifically designed for driving talking faces, AnyTalker achieves the best or competitive results across all metrics. 
Moreover, the 1.3B model of AnyTalker significantly outperforms AniPortrait~\cite{wei2024aniportrait}, EchoMimic~\cite{chen2025echomimic}, and StableAvatar~\cite{tu2025stableavatar} in terms of lip synchronization, even though they have a similar number of parameters. 
These results demonstrate the excellent and comprehensive driving capabilities of the AnyTalker framework.

Subsequently, we evaluate AnyTalker's ability to drive multiple IDs while maintaining both accurate lip synchronization and natural interactivity using the multi-person dataset, InteractiveEyes, described in~\cref{subsec:metric}, along with relevant metrics. 
In this comparison, we contrast AnyTalker with the available open-source multi-person driving methods, MultiTalk~\cite{kong2025let} and Bind-Your-Avatar~\cite{huang2025bind}. 
The results depicted in~\cref{tab:multi-benchmark} demonstrate that both the 1.3B and 14B models of AnyTalker achieve the best performance in terms of the \textbf{Interactivity} metric.
Additionally, the 14B model achieves the best results across all metrics, thereby validating the effectiveness of our proposed training pipeline.  
We further illustrate AnyTalker's capability to generate videos rich in interactivity through quantitative evaluation.

\begin{figure*}[t]
    \centering
    \includegraphics[width=\linewidth]{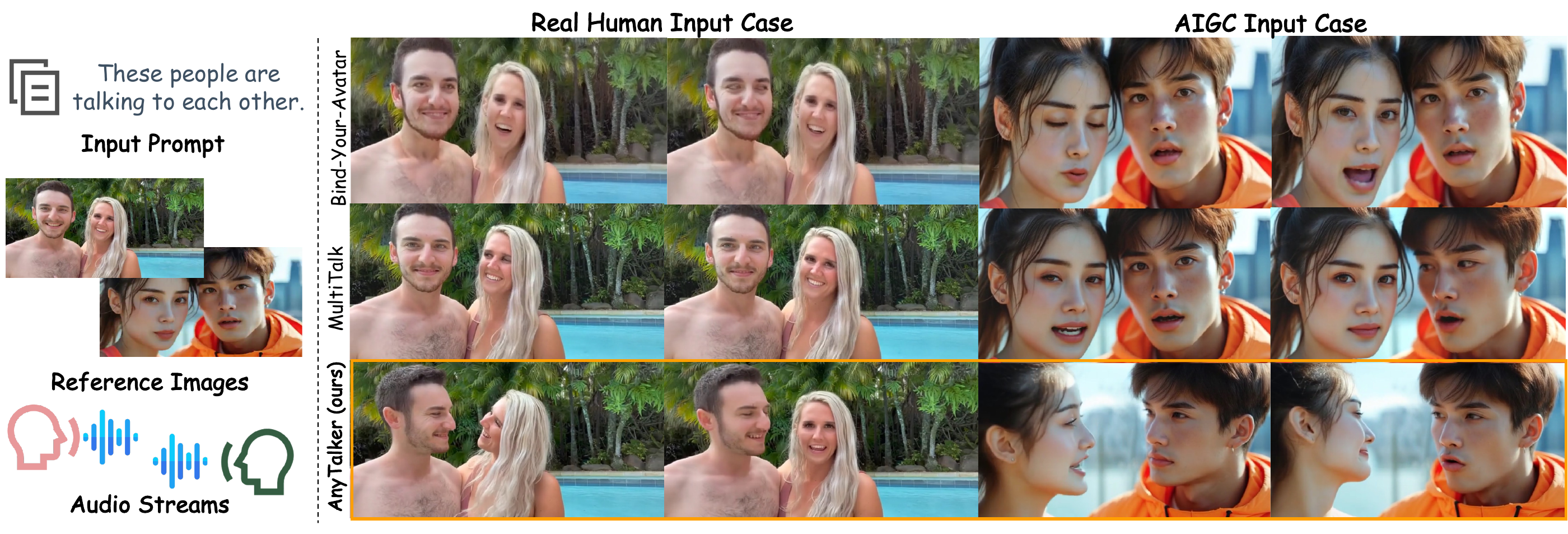}
    \caption{
    Qualitative comparison of multiple multi-person driving methods. 
    With the same text prompt, reference images, and multiple audio streams as input, we compare the generation results of Bind-Your-Avatar, MultiTalk, and AnyTalker. 
    The left case uses the input image from the InteractiveEyes dataset, while the right case uses the image produced by a text-to-image generative model~\cite{kolors}.
    }
    \label{fig:qualitative}
\end{figure*}

\noindent \textbf{Qualitative Comparsion.}
We then select an authentic human input from the InteractiveEyes dataset and use an input generated by an AIGC mode~\cite{kolors}, both accompanied by corresponding text prompts and dual audio streams, to conduct a quantitative evaluation comparison using Bind-Your-Avatar~\cite{huang2025bind}, MultiTalk~\cite{kong2025let}, and AnyTalker. 
As shown in~\cref{fig:qualitative}, AnyTalker generates more natural videos with eye and head interactions compared to the other methods. 
MultiTalk exhibits weaker eye interaction, while Bind-Your-Avatar tends to produce more static expressions. 
This trend further validates the effectiveness of the Interactivity metric proposed in~\cref{subsec:metric}. 
AnyTalker not only generates natural, two-person interactive speaking scenarios but also scales well to multiple IDs, as demonstrated in~\cref{fig:teaser}, where it effectively handles interactions among four IDs.
The qualitative results of the single-person benchmarks~\cite{zhang2021flow, xie2022vfhq} will be included in the supplementary material.

\begin{table}[t]
    \centering
    \caption{
        Ablation study about AnyTalker's components on the HDTF dataset using the 1.3B model.
        ``Baseline'' denotes the model equipped solely with the basic audio attention layers. 
        ``AFCA'' indicates the inclusion of the Audio-Face Cross Attention mechanism. 
        ``Single'' indicates that authentic multi-person data is not utilized in this stage.
    }
    \small
    \begin{tabular}{cccccc}
        \toprule
        \multirow{2}{*}{\textbf{Setting}} & \multicolumn{4}{c}{\textbf{Metrics}} \\
        \cmidrule(lr){2-5}
        & \textbf{Sync-C$\uparrow$} & \textbf{FID$\downarrow$} & \textbf{FVD$\downarrow$} & \textbf{ID$\uparrow$} \\
        \midrule
        Baseline & 5.42 & \textbf{13.01} & \underline{170.96} & \underline{0.90} \\
        w/o AFCA & \underline{6.71} & 14.97 & 207.47 & 0.88 \\
        w/o Mask Token & 5.84 & 14.81 & 193.78 & 0.89 \\
        w/o Concatenated Data & 6.21 & \underline{14.73} & 202.01 & \textbf{0.91} \\
        AnyTalker-1.3B (Single) & \textbf{6.97} & 15.58 & \textbf{166.27} & \textbf{0.91} \\
        \bottomrule
    \end{tabular}
    \label{tab:ab-component}
\end{table}

\subsection{Ablation Study}
\label{subsec:ab}

\noindent \textbf{Components.}
We conduct ablation studies on the three important components mentioned in~\cref{fig:method}, including Audio-Face Cross Attention, Mask Token for attention output, and concatenated multi-person data.
Starting from the full 1.3B model in the first stage, which utilizes only single-person data, we progressively remove the relevant components to evaluate their impact on lip synchronization, generation quality, and identity similarity in the generated videos. 
The results in~\cref{tab:ab-component} show that every component plays a significant role, with concatenated multi-person data most beneficial to lip-movement accuracy.
Although the completed model in the initial phase exhibits a marginally higher FID score compared to the Baseline model, we attribute this to the Baseline model's propensity for generating talking videos with minimal facial expressions or head movements. 
Conversely, the completed model generates more dynamic actions, which inherently influence the FID calculation to some degree. 
We regard this trade-off as a natural consequence of the model's design. Furthermore, the completed model demonstrates superior performance over the Baseline model across other evaluation metrics. 
Crucially, the first-stage model already has the basic capability to drive multiple individuals, a feature that the Baseline model lacks.

\begin{table}[t]
    \centering
    \small
    \caption{
        Ablation studies conducted on the InteractiveEyes dataset using the 1.3B model. 
        ``RS'' denotes the use of authentic single-person data in the first stage. 
        ``CM'' indicates concatenated multi-person data in the first stage. 
        ``RM'' represents authentic multi-person data in the second stage.
    }
    \begin{tabular}{cccccc}
        \toprule
        \multicolumn{3}{c}{\textbf{Setting}} & \multicolumn{3}{c}{\textbf{Metrics}} \\
        \cmidrule(lr){1-3} \cmidrule(lr){4-6}
        \textbf{RS} & \textbf{CM} & \textbf{RM} & \textbf{Interactivity$\uparrow$}& \textbf{Sync-C$^{*}$$\uparrow$} & \textbf{FVD$\downarrow$}  \\
        \midrule
        $\times$ & $\checkmark$ & $\times$ & 0.55 & 3.21 & 672.18  \\
        $\checkmark$ & $\times$ & $\times$ & 0.47 & 4.13 & 475.31  \\
        $\checkmark$ & $\checkmark$ & $\times$ & 0.58 & \textbf{4.89} & \textbf{393.86}  \\
        \midrule
        $\checkmark$ & $\times$ & $\checkmark$ & \underline{0.71} & 3.63 & 511.51 \\
        $\checkmark$ & $\checkmark$ & $\checkmark$ & \textbf{0.97} & \underline{4.56} & \underline{467.84}  \\
        \bottomrule
    \end{tabular}
    \label{tab:ab-data}
\end{table}

\noindent \textbf{Multi-Person Data.}
Subsequently, we focus on multi-person data and conduct more targeted ablation experiments on the InteractiveEyes benchmark. 
In the first stage, we create concatenated multi-person data using single-person data. As shown in~\cref{tab:ab-data}, models utilizing this concatenated data outperform those without it across all evaluation dimensions, especially in lip synchronization and interactivity, highlighting the role of concatenated data in learning multi-person speaking patterns. 
It is worth noting that the results in the first row of the table also demonstrate the necessity of mixing single-person data in the first stage; otherwise, the generated results would be unstable.

After fine-tuning with authentic multi-person data, models that use concatenated data in the first stage demonstrate even more superior performance, further proving their early adaptation to multi-person speaking patterns. 
Although fine-tuning with authentic multi-person data leads to a slight decrease in lip synchronization, it significantly improves interactivity, which we consider a reasonable trade-off.

%% file: sec/6_conclusion.tex
\section{Conclusion}
\label{sec:conclusion}
In this paper, we introduce AnyTalker, an audio-driven framework for generating multi-person talking videos. 
It presents an extensible multi-stream processing structure called Audio-Face Cross Attention that enables identity scaling while guaranteeing seamless cross-identity interactions. 
We further propose a generalizable training strategy that maximally leverages single-person data through concatenation-based augmentation for learning multi-person speaking patterns.
Additionally, we propose the first interactivity evaluation metric and a tailored benchmark for comprehensive assessment. 
Extensive experiments suggest that AnyTalker balances lip synchronization, identity scalability, and interactivity in multi-person scenarios.

%% file: sec/X_suppl.tex
\clearpage
\setcounter{page}{1}
\setcounter{section}{0}
\renewcommand{\thesection}{\Alph{section}} 
\maketitlesupplementary

\section*{Outline}
\label{sec:outline}
This supplementary material provides two sets of additional information on AnyTalker.
\subsection*{A. Experimental Details}
\begin{itemize}
\item Data-processing pipeline used during training
\item Construction scheme for inference data
\item Training hyper-parameters and other technical details
\item Inference settings
\end{itemize}
\subsection*{B. Extended Experiments}
\begin{itemize}
\item More analysis of the Interactivity metric
\item Additional experimental results
\item Effectiveness of interactivity refinement
\end{itemize}

\section*{A. Experimental Details}
\label{sec:data-processing}
\subsection*{A.1. Training Data Processing}
\label{subsec:training-data-processing}
\begin{figure}[t]
    \centering
\includegraphics[width=\linewidth]{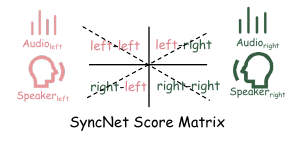}
    \caption{SyncNet Score Matrix in two-person data.}
    \label{fig:sync}
\end{figure}

\noindent \textbf{Auxiliary Models.}
We first introduce the auxiliary models used during the data processing stage.
To segment sentences and chunk the audio track extracted from each video, we employ the pre-trained speaker-diarization-3.1 model~\cite{Plaquet23}\footnote{\url{https://huggingface.co/pyannote/speaker-diarization-3.1}} in conjunction with OpenAI’s Whisper~\cite{radford2023robust}\footnote{\url{https://github.com/openai/whisper}}.
For vocal separation and vocal feature extraction, we use pre-trained models
from Kim\_Vocal\_2~\cite{takahashi2017multi}\footnote{\url{https://github.com/Anjok07/ultimatevocalremovergui}}
and Wav2Vec2~\cite{schneider2019wav2vec,baevski2020wav2vec}\footnote{\url{https://huggingface.co/facebook/wav2vec2-base-960h}}.
For face detection and facial feature processing, we adopt RetinaFace~\cite{deng2020retinaface}
and ArcFace~\cite{deng2019arcface}, both accessible via InsightFace\footnote{\url{https://github.com/deepinsight/insightface}}.
To measure audio-visual synchronization, we employ the pre-trained SyncNet model from LatentSync~\cite{li2024latentsync}\footnote{\url{https://huggingface.co/ByteDance/LatentSync/tree/main}}.
Text prompts are obtained with Gemini~2.5~Pro\footnote{\url{https://deepmind.google/models/gemini/pro/}},
and their features are extracted using the T5-encoder~\cite{raffel2020exploring}\footnote{\url{https://huggingface.co/Wan-AI/Wan2.1-I2V-14B-720P/tree/main/google/umt5-xxl}}.
Finally, we filter videos with excessive camera motion by means of the CoTracker model~\cite{karaev2024cotracker}\footnote{\url{https://huggingface.co/facebook/cotracker}}.

\begin{figure}[t]
    \centering
\includegraphics[width=0.7\linewidth]{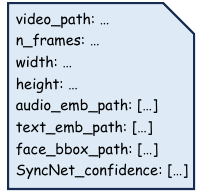}
    \caption{Properties of training data.}
    \label{fig:data}
\end{figure}

\begin{figure*}[t]
    \centering
\includegraphics[width=\linewidth]{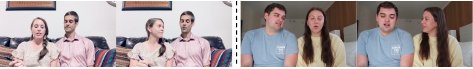}
    \caption{
    Two cases from InteractiveEyes. 
    Both of the speakers have speaking periods.
    }
    \label{fig:multi-data}
\end{figure*}

\begin{figure}[t]
    \centering
\includegraphics[width=\linewidth]{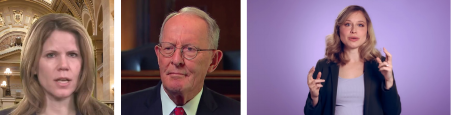}
    \caption{Input cases from the benchmark dataset, from left to right: HDTF~\cite{zhang2021flow}, VFHQ~\cite{xie2022vfhq}, and EMTD~\cite{meng2025echomimicv2}.
    }
    \label{fig:single-data}
\end{figure}

\noindent \textbf{Single-Person Data.}
A valid single-person clip is a 24-fps video that continuously exhibits one identifiable face and is accompanied by speech audio whose lip movements are perfectly synchronised with the soundtrack.
We first eliminate hand-held recordings exhibiting pronounced camera shake or abrupt scene changes using CoTracker~\cite{karaev2024cotracker}.
RetinaFace~\cite{deng2020retinaface} is then applied to guarantee that most frame contains exactly one face.
Videos that satisfy the above criteria are processed by speaker-diarization-3.1~\cite{Plaquet23} to obtain coarse sentence-level segments.
Utterances longer than 5\,s are further subdivided into semantically coherent units with Whisper~\cite{radford2023robust}, after which any fragment containing overlapping speakers is discarded.
The remaining clean single-speaker segments average $\approx$\,2\,s in duration.
Because 2\,s clips cannot saturate GPU memory, we randomly concatenate consecutive segments to extend each sequence; the resulting clip lengths follow a Gaussian distribution with mean 4\,s and standard deviation 0.5\,s.
Then, their vocal tracks are extracted with Kim\_Vocal\_2~\cite{takahashi2017multi} and encoded with Wav2Vec-2~\cite{schneider2019wav2vec}, while audio-visual synchrony is scored by SyncNet~\cite{li2024latentsync}.
Clips whose synchrony score lies below a pre-defined threshold are removed from the training set.
Finally, every retained clip is transcribed and annotated with frame-level face bounding boxes.
We then use Gemini 2.5 Pro to generate textual descriptions for these clips and employ a T5 encoder to extract text features.
All input text is uniformly truncated or padded to exactly 512 tokens.
After being preceded by the above pipeline, it yields approximately $1,000$ h of high-quality 480P single-person data.

\noindent \textbf{Two-Person Data.}
Processing Two-person clips follows the single-person pipeline with three key modifications.
\begin{enumerate}[leftmargin=*,nosep]
  \item \textbf{Face count:} the video must contain \emph{exactly two} faces in most frame.
  \item \textbf{Speaker activity:} speaker-diarization-3.1~\cite{Plaquet23} is constrained to output only two valid states: (i) both speakers active or (ii) a single speaker active. 
  \item \textbf{Spatial consistency:} the left--right spatial ordering of the two faces must remain unchanged throughout the entire clip; identity swapping is detected and rejected via InsightFace~\cite{deng2020retinaface}.
\end{enumerate}
To establish the correct voice--face correspondence, we compute a $2\!\times\!2$ SyncNet confidence matrix (\cref{fig:sync}) and requires the two largest scores to lie on the diagonal; otherwise, the clip is discarded.
A minimum synchrony threshold identical to the single-person setting is further applied.
After filtering, approximately $12$ h of clean two-person data are retained.

\noindent \textbf{Data Concatenation.}
During the first-stage training, we horizontally concatenate randomly selected single-person clips.  
Because the original videos are extremely high-resolution (many 2K/4K), naively resizing them along the height axis and stacking would yield faces that occupy only a tiny fraction of the frame.  
To avoid this, we adopt a special cropping strategy.  
First, we locate the face centre in each of the two candidate clips.  
For 480P footage, the frame size is (H, W)=(480,832), so we expand a minimal crop of (480,416) around each face centre.  
We then apply further augmentation: while keeping the crop inside the original image, we randomly enlarge the window while preserving the 480/416 aspect ratio.  
The resulting crops contain a much larger facial region, enabling the model to learn a more accurate lip-to-audio mapping.

\noindent \textbf{Data Properties.}
Each processed sample is summarised by a dictionary-style item that stores the absolute paths to the decoded video, cleaned audio, and pre-extracted features.
During training, the data loader indexes all information through this item.
Single-person and two-person clips share an identical key structure; the number of face--speech entries in the list unambiguously indicates whether the sample contains one or two speakers.
Further details are illustrated in~\cref{fig:data}.

\begin{figure*}[t]
    \centering
\includegraphics[width=0.9\linewidth]{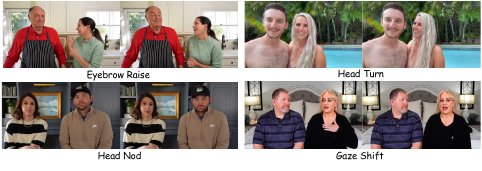}
    \caption{
    Four typical cases that will get a high Interactivity score.
    }
    \label{fig:case}
\end{figure*}

\subsection*{A.2. Benchmark Data Processing}
\label{subsec:benchmark-data-processing}

\noindent \textbf{Single-Person Data.}
We adopt HDTF~\cite{zhang2021flow} and VFHQ~\cite{xie2022vfhq} as the single-speaker evaluation benchmarks.
Both datasets are de facto standards for talking-head generation; their images are tightly cropped around the face region.
Additional experiments on half-body sequences that include hands are reported in Sec.~B.2.

For each test video, we supply (i) a single reference image of the subject and (ii) the corresponding speech segment.
To respect GPU-memory constraints, we fix the audio length to 6\,s---a duration that all competing methods process without overflow.
Twenty clips are randomly selected from each dataset, ensuring that none of the identities appear in the AnyTalker training set.
Reference images from the three evaluation splits are visualised in~\cref{fig:single-data}.

\noindent \textbf{Two-Person Data.}
We provide additional details for \emph{InteractiveEyes}, the test set introduced in Section 4 in the main text for two-speaker scenarios.
Each video is shot with a stable camera and contains \emph{exactly two faces in every frame}; the duration is approximately 10\,s.
In 80\,\% of the cases, both speakers produce speech, while in the remaining 20\,\% only one speaker talks and the other maintains a listening pose.
To avoid segmentation errors, we do \emph{not} apply speaker-diarization-3.1~\cite{Plaquet23}; instead, speaking intervals are manually annotated so that the Interactivity metric proposed in the main paper can be computed accurately.
Two representative cases are shown in~\cref{fig:multi-data}.

\begin{table}[t]
    \centering
    \caption{
       Quantitative result on EMTD~\cite{meng2025echomimicv2} benchmark.
    }
    \small
    \begin{tabular}{cccccc}
        \toprule
        \multirow{2}{*}{\textbf{Method}} & \multicolumn{4}{c}{\textbf{Metrics}} \\
        \cmidrule(lr){2-5}
        & \textbf{Sync-C$\uparrow$} & \textbf{FID$\downarrow$} & \textbf{FVD$\downarrow$} & \textbf{ID$\uparrow$} \\
        \midrule
        FantasyTalking~\cite{wang2025fantasytalking} & 3.76 & 67.66 & 818.71 & 0.76 \\
        EchoMimic v2~\cite{meng2025echomimicv2} & 6.27 & 63.43 & \underline{671.18} & 0.76 \\
        MultiTalk~\cite{kong2025let} & \underline{8.38} & 64.71 & 787.99 & \textbf{0.79} \\
        \midrule
        AnyTalker-1.3B & 5.83 & \underline{56.01} & 789.59 & 0.74 \\
        AnyTalker-14B & \textbf{8.45} & \textbf{50.61} & \textbf{664.58} & \underline{0.77} \\
        \bottomrule
    \end{tabular}
    \label{tab:EMTD}
\end{table}

\begin{figure*}[t]
    \centering
\includegraphics[width=\linewidth]{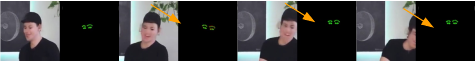}
    \caption{
   A bad case generated by Bind-Your-Avatar~\cite{huang2025bind}.
    }
    \label{fig:bad}
\end{figure*}

\begin{figure*}[t]
    \centering
\includegraphics[width=0.9\linewidth]{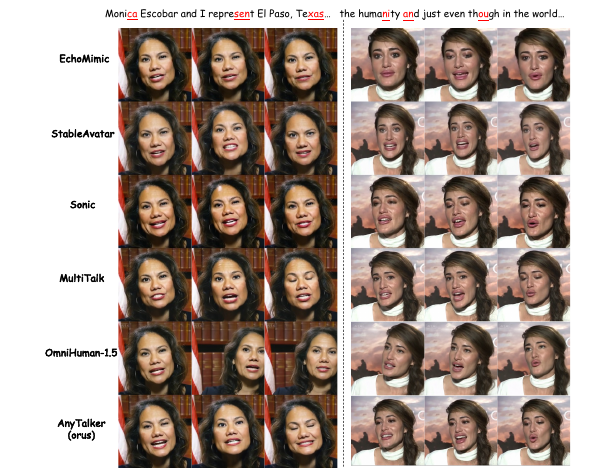}
    \caption{
   Qualitative results on HDTF~\cite{zhang2021flow} (left) and VFHQ~\cite{xie2022vfhq} (right) benchmark.
   The positions of pronunciation have been highlighted in {\color{red}red} and \underline{underlined}.
    }
    \label{fig:qualitative-single}
\end{figure*}

\subsection*{A.3. Implement Details}
\label{sec:exp-detail}

\noindent \textbf{Training Details.}
To comprehensively evaluate our method, we train two models of different sizes: Wan2.1-1.3B-Inp\footnote{\url{https://huggingface.co/alibaba-pai/Wan2.1-Fun-1.3B-InP}} and Wan2.1-I2V-14B~\cite{wan2025wan}, which served as the foundational video diffusion models for our experiments. 
In all stages, the text~\cite{raffel2020exploring}, audio~\cite{baevski2020wav2vec}, and image~\cite{radford2021learning} encoders, as well as the 3D VAE, remained frozen with their parameters unchanged. 
The DiT main network, including the newly added AFCA Layers, had all its parameters open for training. 
The first stage employs a higher learning rate of \(2 \times 10^{-5}\) for pretraining, while the second stage uses a lower learning rate of \(5 \times 10^{-6}\) for fine-tuning, incorporating a warm-up strategy and optimized using the AdamW optimizer~\cite{loshchilov2017decoupled}. 
The 14B model is trained using 32 NVIDIA H200 GPUs. 
In the first stage, the global batch size is set to 32 and the model is trained for 2.4M steps. 
In the second stage, the batch size is adjusted to 16, and the model is trained for an additional 50K steps.
The 1.3B model is trained using 8 NVIDIA H200 GPUs. 
The global batch size is maintained at 48 throughout the training process, with the total number of training steps being consistent with that of the 14B model.

\noindent \textbf{Inference Details.}
For all competing methods, we strictly follow the publicly released implementations and adopt their default recommended inference hyperparameters.
Approaches that require textual input receive a fixed prompt for the single-person benchmark: ``\texttt{this person is talking}''.
For the multi-person benchmark, the prompts are automatically generated by Gemini~2.5~Pro as described in~Sec.~A.1.
AnyTalker performs inference with classifier-free guidance (CFG)~\citep{ho2021classifier} using a guidance scale of $4.0$; in the unconditional branch, both textual and audio features are set to zero.
Face masks are extracted with InsightFace and uniformly dilated to provide a slightly enlarged region for generation.

\section*{B. Extended Experiments}
\label{sec:exp-res}

\subsection*{B.1. More analysis about Interactivity Metric}
\label{subsec:S-metric}

\begin{figure*}[t]
    \centering
\includegraphics[width=\linewidth]{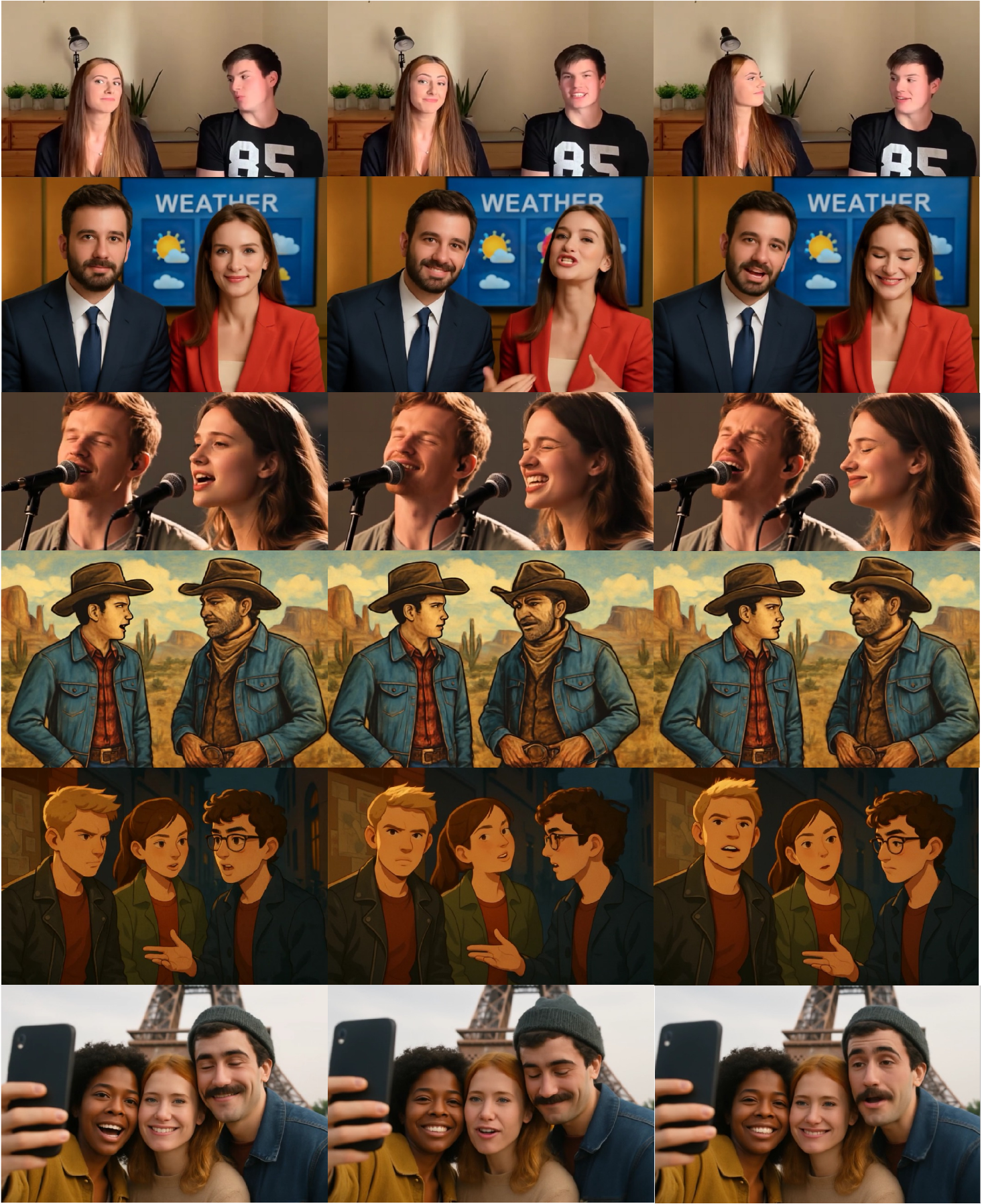}
    \caption{
   More Results generated by AnyTalker.
    }
    \label{fig:anytalker-result}
\end{figure*}

\begin{figure*}[t]
    \centering
\includegraphics[width=0.6\linewidth]{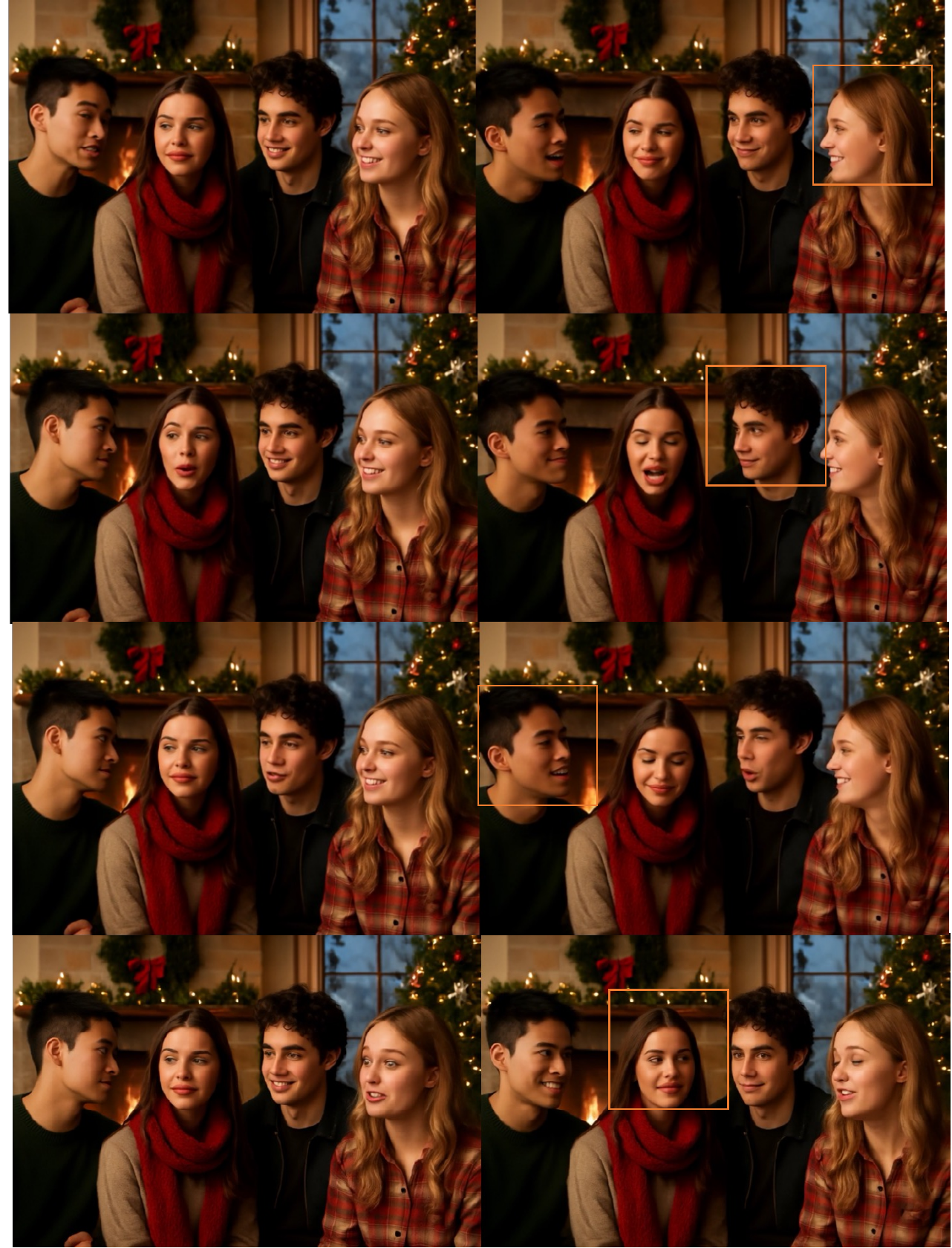}
    \caption{
   Improvement in interaction among each identity after fine-tuning on authentic multi-person data (right). 
    }
    \label{fig:ab-inter}
\end{figure*}

\noindent \textbf{Good Case.}
\cref{fig:case} visualises four representative listener behaviours that strongly signal conversational engagement: eyebrow raise, head nod, head turn, and gaze shift.
The presence of such cues contributes substantially to the final Interactivity score.
Because Interactivity is computed over the entire video, we do not report frame-level values in~\cref{fig:case}.
As illustrated in~\cref{fig:anytalker-result}, AnyTalker produces sequences rich in these interactive actions, and thus achieves a comparatively high Interactivity rating.

\noindent \textbf{The Rubustness of Interactive Metric.}
Some generative baselines occasionally produce highly implausible motions.
For example, Bind-Your-Avatar~\citep{huang2025bind} generates an exaggerated ``lying-down'' action shown in~\cref{fig:bad}.
Without counter-measures, such artifacts would dramatically inflate the ${Motion}$ score even though they are unrelated to interactivity.
We therefore introduce a lightweight anomaly-suppression rule: if the mean facial-landmark displacement between two consecutive frames exceeds 10 pixels (all faces are pre-aligned to a $256\times 256$ canvas), the landmark positions are frozen until a subsequent displacement below 10 pixels is observed.
As demonstrated in the right two frames of~\cref{fig:bad}, this simple clamping prevents the vast majority of abnormal movements from entering the Interactivity computation.

\subsection*{B.2. Additional Experimental Results}
\label{subsec:more-quali}

\noindent \textbf{Quantitative Results on the EMTD Benchmark.}
EMTD~\citep{meng2025echomimicv2} is a half-body dataset that includes hands, as illustrated in~\cref{fig:data}.
We compare AnyTalker against three methods capable of generating half-body sequences: EchoMimic\,v2~\citep{meng2025echomimicv2}, FantasyTalking~\citep{wang2025fantasytalking}, and MultiTalk~\citep{kong2025let}.
The evaluation protocol strictly follows the metrics described in Sec.\,5.1 for the single-person benchmark.
AnyTalker-14B achieves the best scores on three metrics and is only marginally behind MultiTalk on identity preservation (ID).
These results demonstrate that AnyTalker is a comprehensive framework that handles not only tight-face inputs but also half-body scenarios.

\noindent \textbf{Qualitative Result on Single-Person Benchmarks.}
As illustrated in~\cref{fig:qualitative-single}, we present qualitative comparisons of EchoMimic~\cite{chen2025echomimic}, StableAvatar~\cite{tu2025stableavatar}, Sonic~\cite{ji2025sonic}, MultiTalk~\cite{kong2025let}, OmniHuman-1.5~\cite{jiang2025omnihuman}, and AnyTalker-14B on the HDTF~\cite{zhang2021flow} and VFHQ~\cite{xie2022vfhq} benchmarks.
AnyTalker consistently produces clear dentition and accurate lip movement.

\noindent \textbf{More Multi-Person Results.}
\cref{fig:anytalker-result} showcases additional multi-person animations generated by AnyTalker. The model gracefully handles a broad spectrum of inputs: real photographs, AIGC images, and cartoons. 
It produces natural, context-appropriate interactions regardless of the number of identities involved. Further compelling examples are available in the videos on the project homepage.

\subsection*{B.3. Effectiveness of Interactivity Refinement}
\label{sec:ab}
As shown in~\cref{fig:ab-inter}, refining interactivity with authentic multi-person data lets identities exchange significantly more natural eye contact.
A model trained solely on single-person data can activate each face correctly, yet every identity remains blank whenever it is not speaking, an effect that looks highly unnatural in conversational scenes.

\subsection*{B.4. Future Work}
At present, AnyTalker supports only rudimentary camera motions driven by textual prompts.
Drawing inspiration from recent controllable video-generation approaches~\cite{ye2025unic, lu2024coarse, ye2025stylemaster}, we can incorporate additional conditional signals, such as camera trajectory~\cite {bai2025recammaster}.
By grafting lightweight, training-efficient modules~\cite{hu2022lora,kong2025taming,lu2025adversarial,ren2024hyper} into recent camera-trajectory-control techniques~\cite{zhang2025zero,bahmani2025ac3d}, we expect to enrich the visual storytelling of the generated videos, automatically framing and tracking the active speaker without manual intervention.

%% file: main.bib
@String(CVPR= {IEEE Conf. Comput. Vis. Pattern Recog.})

@String(ICLR = {Int. Conf. Learn. Represent.})

@String(AAAI = {AAAI})

@String(CVPR  = {CVPR})

@String(ICLR  = {ICLR})

@article{blattmann2023stable,
  title={Stable video diffusion: Scaling latent video diffusion models to large datasets},
  author={Blattmann, Andreas and Dockhorn, Tim and Kulal, Sumith and Mendelevitch, Daniel and Kilian, Maciej and Lorenz, Dominik and Levi, Yam and English, Zion and Voleti, Vikram and Letts, Adam and others},
  journal={arXiv preprint arXiv:2311.15127},
  year={2023}
}

@inproceedings{yangcogvideox,
  title={CogVideoX: Text-to-Video Diffusion Models with An Expert Transformer},
  author={Yang, Zhuoyi and Teng, Jiayan and Zheng, Wendi and Ding, Ming and Huang, Shiyu and Xu, Jiazheng and Yang, Yuanming and Hong, Wenyi and Zhang, Xiaohan and Feng, Guanyu and others},
  booktitle={The Thirteenth International Conference on Learning Representations},
  year={2025}
}

@article{kong2024hunyuanvideo,
  title={Hunyuanvideo: A systematic framework for large video generative models},
  author={Kong, Weijie and Tian, Qi and Zhang, Zijian and Min, Rox and Dai, Zuozhuo and Zhou, Jin and Xiong, Jiangfeng and Li, Xin and Wu, Bo and Zhang, Jianwei and others},
  journal={arXiv preprint arXiv:2412.03603},
  year={2024}
}

@article{wan2025wan,
  title={Wan: Open and advanced large-scale video generative models},
  author={Wan, Team and Wang, Ang and Ai, Baole and Wen, Bin and Mao, Chaojie and Xie, Chen-Wei and Chen, Di and Yu, Feiwu and Zhao, Haiming and Yang, Jianxiao and others},
  journal={arXiv preprint arXiv:2503.20314},
  year={2025}
}

@inproceedings{lin2025omnihuman,
  title={Omnihuman-1: Rethinking the scaling-up of one-stage conditioned human animation models},
  author={Lin, Gaojie and Jiang, Jianwen and Yang, Jiaqi and Zheng, Zerong and Liang, Chao and Zhang, Yuan and Liu, Jingtuo},
  booktitle={Proceedings of the IEEE/CVF International Conference on Computer Vision},
  pages={13847--13858},
  year={2025}
}

@article{jiang2025omnihuman,
  title={Omnihuman-1.5: Instilling an active mind in avatars via cognitive simulation},
  author={Jiang, Jianwen and Zeng, Weihong and Zheng, Zerong and Yang, Jiaqi and Liang, Chao and Liao, Wang and Liang, Han and Zhang, Yuan and Gao, Mingyuan},
  journal={arXiv preprint arXiv:2508.19209},
  year={2025}
}

@inproceedings{cui2025hallo3,
  title={Hallo3: Highly dynamic and realistic portrait image animation with video diffusion transformer},
  author={Cui, Jiahao and Li, Hui and Zhan, Yun and Shang, Hanlin and Cheng, Kaihui and Ma, Yuqi and Mu, Shan and Zhou, Hang and Wang, Jingdong and Zhu, Siyu},
  booktitle={Proceedings of the Computer Vision and Pattern Recognition Conference},
  pages={21086--21095},
  year={2025}
}

@article{kong2025let,
  title={Let Them Talk: Audio-Driven Multi-Person Conversational Video Generation},
  author={Kong, Zhe and Gao, Feng and Zhang, Yong and Kang, Zhuoliang and Wei, Xiaoming and Cai, Xunliang and Chen, Guanying and Luo, Wenhan},
  journal={arXiv preprint arXiv:2505.22647},
  year={2025}
}

@article{yang2025infinitetalk,
  title={InfiniteTalk: Audio-driven Video Generation for Sparse-Frame Video Dubbing},
  author={Yang, Shaoshu and Kong, Zhe and Gao, Feng and Cheng, Meng and Liu, Xiangyu and Zhang, Yong and Kang, Zhuoliang and Luo, Wenhan and Cai, Xunliang and He, Ran and others},
  journal={arXiv preprint arXiv:2508.14033},
  year={2025}
}

@inproceedings{ji2025sonic,
  title={Sonic: Shifting focus to global audio perception in portrait animation},
  author={Ji, Xiaozhong and Hu, Xiaobin and Xu, Zhihong and Zhu, Junwei and Lin, Chuming and He, Qingdong and Zhang, Jiangning and Luo, Donghao and Chen, Yi and Lin, Qin and others},
  booktitle={Proceedings of the Computer Vision and Pattern Recognition Conference},
  pages={193--203},
  year={2025}
}

@article{gan2025omniavatar,
  title={OmniAvatar: Efficient Audio-Driven Avatar Video Generation with Adaptive Body Animation},
  author={Gan, Qijun and Yang, Ruizi and Zhu, Jianke and Xue, Shaofei and Hoi, Steven},
  journal={arXiv preprint arXiv:2506.18866},
  year={2025}
}

@article{chen2025hunyuanvideo,
  title={HunyuanVideo-Avatar: High-Fidelity Audio-Driven Human Animation for Multiple Characters},
  author={Chen, Yi and Liang, Sen and Zhou, Zixiang and Huang, Ziyao and Ma, Yifeng and Tang, Junshu and Lin, Qin and Zhou, Yuan and Lu, Qinglin},
  journal={arXiv preprint arXiv:2505.20156},
  year={2025}
}

@inproceedings{jiangloopy,
  title={Loopy: Taming Audio-Driven Portrait Avatar with Long-Term Motion Dependency},
  author={Jiang, Jianwen and Liang, Chao and Yang, Jiaqi and Lin, Gaojie and Zhong, Tianyun and Zheng, Yanbo},
  booktitle={The Thirteenth International Conference on Learning Representations},
  year={2025}
}

@inproceedings{tian2024emo,
  title={Emo: Emote portrait alive generating expressive portrait videos with audio2video diffusion model under weak conditions},
  author={Tian, Linrui and Wang, Qi and Zhang, Bang and Bo, Liefeng},
  booktitle={European Conference on Computer Vision},
  pages={244--260},
  year={2024},
  organization={Springer}
}

@inproceedings{chen2025echomimic,
  title={Echomimic: Lifelike audio-driven portrait animations through editable landmark conditions},
  author={Chen, Zhiyuan and Cao, Jiajiong and Chen, Zhiquan and Li, Yuming and Ma, Chenguang},
  booktitle={Proceedings of the AAAI Conference on Artificial Intelligence},
  volume={39},
  number={3},
  pages={2403--2410},
  year={2025}
}

@article{xu2024hallo,
  title={Hallo: Hierarchical audio-driven visual synthesis for portrait image animation},
  author={Xu, Mingwang and Li, Hui and Su, Qingkun and Shang, Hanlin and Zhang, Liwei and Liu, Ce and Wang, Jingdong and Yao, Yao and Zhu, Siyu},
  journal={arXiv preprint arXiv:2406.08801},
  year={2024}
}

@article{wang2024v,
  title={V-express: Conditional dropout for progressive training of portrait video generation},
  author={Wang, Cong and Tian, Kuan and Zhang, Jun and Guan, Yonghang and Luo, Feng and Shen, Fei and Jiang, Zhiwei and Gu, Qing and Han, Xiao and Yang, Wei},
  journal={arXiv preprint arXiv:2406.02511},
  year={2024}
}

@article{gao2025wan,
  title={Wan-s2v: Audio-driven cinematic video generation},
  author={Gao, Xin and Hu, Li and Hu, Siqi and Huang, Mingyang and Ji, Chaonan and Meng, Dechao and Qi, Jinwei and Qiao, Penchong and Shen, Zhen and Song, Yafei and others},
  journal={arXiv preprint arXiv:2508.18621},
  year={2025}
}

@article{nazarieh2024portraittalk,
  title={PortraitTalk: Towards customizable one-shot audio-to-talking face generation},
  author={Nazarieh, Fatemeh and Feng, Zhenhua and Kanojia, Diptesh and Awais, Muhammad and Kittler, Josef},
  journal={arXiv preprint arXiv:2412.07754},
  year={2024}
}

@article{yee2025synchrorama,
  title={SynchroRaMa: Lip-Synchronized and Emotion-Aware Talking Face Generation via Multi-Modal Emotion Embedding},
  author={Yee, Phyo Thet and Kollias, Dimitrios and Mishra, Sudeepta and Dhall, Abhinav},
  journal={arXiv preprint arXiv:2509.19965},
  year={2025}
}

@inproceedings{rombach2022high,
  title={High-resolution image synthesis with latent diffusion models},
  author={Rombach, Robin and Blattmann, Andreas and Lorenz, Dominik and Esser, Patrick and Ommer, Bj{\"o}rn},
  booktitle={Proceedings of the IEEE/CVF Conference on Computer Vision and Pattern Recognition},
  pages={10684--10695},
  year={2022}
}

@inproceedings{peebles2023scalable,
  title={Scalable Diffusion Models with Transformers},
  author={Peebles, William and Xie, Saining},
  booktitle={Proceedings of the IEEE/CVF international conference on computer vision},
  pages={4195--4205},
  year={2023}
}

@inproceedings{guoanimatediff,
  title={AnimateDiff: Animate Your Personalized Text-to-Image Diffusion Models without Specific Tuning},
  author={Guo, Yuwei and Yang, Ceyuan and Rao, Anyi and Liang, Zhengyang and Wang, Yaohui and Qiao, Yu and Agrawala, Maneesh and Lin, Dahua and Dai, Bo},
  booktitle={The Twelfth International Conference on Learning Representations},
  year={2024}
}

@inproceedings{hu2024animate,
  title={Animate anyone: Consistent and controllable image-to-video synthesis for character animation},
  author={Hu, Li},
  booktitle={Proceedings of the IEEE/CVF Conference on Computer Vision and Pattern Recognition},
  pages={8153--8163},
  year={2024}
}

@article{guo2024liveportrait,
  title={Liveportrait: Efficient portrait animation with stitching and retargeting control},
  author={Guo, Jianzhu and Zhang, Dingyun and Liu, Xiaoqiang and Zhong, Zhizhou and Zhang, Yuan and Wan, Pengfei and Zhang, Di},
  journal={arXiv preprint arXiv:2407.03168},
  year={2024}
}

@inproceedings{zhang2023sadtalker,
  title={Sadtalker: Learning realistic 3d motion coefficients for stylized audio-driven single image talking face animation},
  author={Zhang, Wenxuan and Cun, Xiaodong and Wang, Xuan and Zhang, Yong and Shen, Xi and Guo, Yu and Shan, Ying and Wang, Fei},
  booktitle={Proceedings of the IEEE/CVF Conference on Computer Vision and Pattern Recognition},
  pages={8652--8661},
  year={2023}
}

@article{xu2024vasa,
  title={Vasa-1: Lifelike audio-driven talking faces generated in real time},
  author={Xu, Sicheng and Chen, Guojun and Guo, Yu-Xiao and Yang, Jiaolong and Li, Chong and Zang, Zhenyu and Zhang, Yizhong and Tong, Xin and Guo, Baining},
  journal={Advances in Neural Information Processing Systems},
  volume={37},
  pages={660--684},
  year={2024}
}

@article{zhang2024musetalk,
  title={MuseTalk: Real-Time High-Fidelity Video Dubbing via Spatio-Temporal Sampling},
  author={Zhang, Yue and Zhong, Zhizhou and Liu, Minhao and Chen, Zhaokang and Wu, Bin and Zeng, Yubin and Zhan, Chao and He, Yingjie and Huang, Junxin and Zhou, Wenjiang},
  journal={arXiv preprint arXiv:2410.10122},
  year={2024}
}

@article{wang2025fantasyportrait,
  title={FantasyPortrait: Enhancing Multi-Character Portrait Animation with Expression-Augmented Diffusion Transformers},
  author={Wang, Qiang and Wang, Mengchao and Jiang, Fan and Fan, Yaqi and Qi, Yonggang and Xu, Mu},
  journal={arXiv preprint arXiv:2507.12956},
  year={2025}
}

@article{huang2025bind,
  title={Bind-Your-Avatar: Multi-Talking-Character Video Generation with Dynamic 3D-mask-based Embedding Router},
  author={Huang, Yubo and Wang, Weiqiang and Zhao, Sirui and Xu, Tong and Liu, Lin and Chen, Enhong},
  journal={arXiv preprint arXiv:2506.19833},
  year={2025}
}

@article{ma2025playmate2,
  title={Playmate2: Training-Free Multi-Character Audio-Driven Animation via Diffusion Transformer with Reward Feedback},
  author={Ma, Xingpei and Huang, Shenneng and Cai, Jiaran and Guan, Yuansheng and Zheng, Shen and Zhao, Hanfeng and Zhang, Qiang and Zhang, Shunsi},
  journal={arXiv preprint arXiv:2510.12089},
  year={2025}
}

@article{meng2025identity,
  title={Identity-GRPO: Optimizing Multi-Human Identity-preserving Video Generation via Reinforcement Learning},
  author={Meng, Xiangyu and Zhang, Zixian and Zhang, Zhenghao and Liao, Junchao and Qin, Long and Wang, Weizhi},
  journal={arXiv preprint arXiv:2510.14256},
  year={2025}
}

@article{wang2025interacthuman,
  title={InterActHuman: Multi-Concept Human Animation with Layout-Aligned Audio Conditions},
  author={Wang, Zhenzhi and Yang, Jiaqi and Jiang, Jianwen and Liang, Chao and Lin, Gaojie and Zheng, Zerong and Yang, Ceyuan and Lin, Dahua},
  journal={arXiv preprint arXiv:2506.09984},
  year={2025}
}

@inproceedings{zhang2021flow,
  title={Flow-guided one-shot talking face generation with a high-resolution audio-visual dataset},
  author={Zhang, Zhimeng and Li, Lincheng and Ding, Yu and Fan, Changjie},
  booktitle={Proceedings of the IEEE/CVF Conference on Computer Vision and Pattern Recognition},
  pages={3661--3670},
  year={2021}
}

@inproceedings{xie2022vfhq,
  title={Vfhq: A high-quality dataset and benchmark for video face super-resolution},
  author={Xie, Liangbin and Wang, Xintao and Zhang, Honglun and Dong, Chao and Shan, Ying},
  booktitle={Proceedings of the IEEE/CVF Conference on Computer Vision and Pattern Recognition},
  pages={657--666},
  year={2022}
}

@inproceedings{zhu2022celebv,
  title={CelebV-HQ: A large-scale video facial attributes dataset},
  author={Zhu, Hao and Wu, Wayne and Zhu, Wentao and Jiang, Liming and Tang, Siwei and Zhang, Li and Liu, Ziwei and Loy, Chen Change},
  booktitle={European Conference on Computer Vision},
  pages={650--667},
  year={2022},
  organization={Springer}
}

@inproceedings{meng2025echomimicv2,
  title={Echomimicv2: Towards striking, simplified, and semi-body human animation},
  author={Meng, Rang and Zhang, Xingyu and Li, Yuming and Ma, Chenguang},
  booktitle={Proceedings of the Computer Vision and Pattern Recognition Conference},
  pages={5489--5498},
  year={2025}
}

@inproceedings{li2025openhumanvid,
  title={Openhumanvid: A large-scale high-quality dataset for enhancing human-centric video generation},
  author={Li, Hui and Xu, Mingwang and Zhan, Yun and Mu, Shan and Li, Jiaye and Cheng, Kaihui and Chen, Yuxuan and Chen, Tan and Ye, Mao and Wang, Jingdong and others},
  booktitle={Proceedings of the Computer Vision and Pattern Recognition Conference},
  pages={7752--7762},
  year={2025}
}

@article{chung2018voxceleb2,
  title={VoxCeleb2: Deep speaker recognition},
  author={Chung, J and Nagrani, A and Zisserman, A},
  journal={Interspeech 2018},
  year={2018},
  publisher={International Speech Communication Association}
}

@inproceedings{xuetowards,
  title={Towards Multiple Character Image Animation Through Enhancing Implicit Decoupling},
  author={Xue, Jingyun and Wang, Hongfa and Tian, Qi and Ma, Yue and Wang, Andong and Zhao, Zhiyuan and Min, Shaobo and Zhao, Wenzhe and Zhang, Kaihao and Shum, Heung-Yeung and others},
  booktitle={The Thirteenth International Conference on Learning Representations},
  year={2025}
}

@article{ho2020denoising,
  title={Denoising diffusion probabilistic models},
  author={Ho, Jonathan and Jain, Ajay and Abbeel, Pieter},
  journal={Advances in Neural Information Processing Systems},
  volume={33},
  pages={6840--6851},
  year={2020}
}

@inproceedings{ho2021classifier,
  title={Classifier-Free Diffusion Guidance},
  author={Ho, Jonathan and Salimans, Tim},
  booktitle={NeurIPS 2021 Workshop on Deep Generative Models and Downstream Applications},
  year={2021}
}

@inproceedings{zhang2023adding,
  title={Adding conditional control to text-to-image diffusion models},
  author={Zhang, Lvmin and Rao, Anyi and Agrawala, Maneesh},
  booktitle={Proceedings of the IEEE/CVF international conference on computer vision},
  pages={3836--3847},
  year={2023}
}

@inproceedings{lin2025cyberhost,
  title={Cyberhost: A one-stage diffusion framework for audio-driven talking body generation},
  author={Lin, Gaojie and Jiang, Jianwen and Liang, Chao and Zhong, Tianyun and Yang, Jiaqi and Zheng, Zerong and Zheng, Yanbo},
  booktitle={The Thirteenth International Conference on Learning Representations},
  year={2025}
}

@inproceedings{zhu2023tryondiffusion,
  title={Tryondiffusion: A tale of two unets},
  author={Zhu, Luyang and Yang, Dawei and Zhu, Tyler and Reda, Fitsum and Chan, William and Saharia, Chitwan and Norouzi, Mohammad and Kemelmacher-Shlizerman, Ira},
  booktitle={Proceedings of the IEEE/CVF Conference on Computer Vision and Pattern Recognition},
  pages={4606--4615},
  year={2023}
}

@inproceedings{radford2023robust,
  title={Robust speech recognition via large-scale weak supervision},
  author={Radford, Alec and Kim, Jong Wook and Xu, Tao and Brockman, Greg and McLeavey, Christine and Sutskever, Ilya},
  booktitle={International conference on machine learning},
  pages={28492--28518},
  year={2023},
  organization={PMLR}
}

@article{baevski2020wav2vec,
  title={wav2vec 2.0: A framework for self-supervised learning of speech representations},
  author={Baevski, Alexei and Zhou, Yuhao and Mohamed, Abdelrahman and Auli, Michael},
  journal={Advances in Neural Information Processing Systems},
  volume={33},
  pages={12449--12460},
  year={2020}
}

@article{schneider2019wav2vec,
  title={wav2vec: Unsupervised pre-training for speech recognition},
  author={Schneider, Steffen and Baevski, Alexei and Collobert, Ronan and Auli, Michael},
  journal={arXiv preprint arXiv:1904.05862},
  year={2019}
}

@article{vaswani2017attention,
  title={Attention is all you need},
  author={Vaswani, Ashish and Shazeer, Noam and Parmar, Niki and Uszkoreit, Jakob and Jones, Llion and Gomez, Aidan N and Kaiser, {\L}ukasz and Polosukhin, Illia},
  journal={Advances in Neural Information Processing Systems},
  volume={30},
  year={2017}
}

@article{meng2025echomimicv3,
  title={Echomimicv3: 1.3 b parameters are all you need for unified multi-modal and multi-task human animation},
  author={Meng, Rang and Wang, Yan and Wu, Weipeng and Zheng, Ruobing and Li, Yuming and Ma, Chenguang},
  journal={arXiv preprint arXiv:2507.03905},
  year={2025}
}

@article{tu2025stableavatar,
  title={Stableavatar: Infinite-length audio-driven avatar video generation},
  author={Tu, Shuyuan and Pan, Yueming and Huang, Yinming and Han, Xintong and Xing, Zhen and Dai, Qi and Luo, Chong and Wu, Zuxuan and Jiang, Yu-Gang},
  journal={arXiv preprint arXiv:2508.08248},
  year={2025}
}

@article{su2024roformer,
  title={Roformer: Enhanced transformer with rotary position embedding},
  author={Su, Jianlin and Ahmed, Murtadha and Lu, Yu and Pan, Shengfeng and Bo, Wen and Liu, Yunfeng},
  journal={Neurocomputing},
  volume={568},
  pages={127063},
  year={2024},
  publisher={Elsevier}
}

@article{chen2025midas,
  title={Midas: Multimodal interactive digital-human synthesis via real-time autoregressive video generation},
  author={Chen, Ming and Cui, Liyuan and Zhang, Wenyuan and Zhang, Haoxian and Zhou, Yan and Li, Xiaohan and Tang, Songlin and Liu, Jiwen and Liao, Borui and Chen, Hejia and others},
  journal={arXiv preprint arXiv:2508.19320},
  year={2025}
}

@inproceedings{radford2021learning,
  title={Learning transferable visual models from natural language supervision},
  author={Radford, Alec and Kim, Jong Wook and Hallacy, Chris and Ramesh, Aditya and Goh, Gabriel and Agarwal, Sandhini and Sastry, Girish and Askell, Amanda and Mishkin, Pamela and Clark, Jack and others},
  booktitle={International conference on machine learning},
  pages={8748--8763},
  year={2021},
  organization={PmLR}
}

@article{raffel2020exploring,
  title={Exploring the limits of transfer learning with a unified text-to-text transformer},
  author={Raffel, Colin and Shazeer, Noam and Roberts, Adam and Lee, Katherine and Narang, Sharan and Matena, Michael and Zhou, Yanqi and Li, Wei and Liu, Peter J},
  journal={Journal of machine learning research},
  volume={21},
  number={140},
  pages={1--67},
  year={2020}
}

@inproceedings{deng2020retinaface,
  title={Retinaface: Single-shot multi-level face localisation in the wild},
  author={Deng, Jiankang and Guo, Jia and Ververas, Evangelos and Kotsia, Irene and Zafeiriou, Stefanos},
  booktitle={Proceedings of the IEEE/CVF Conference on Computer Vision and Pattern Recognition},
  pages={5203--5212},
  year={2020}
}

@inproceedings{Plaquet23,
  author={Alexis Plaquet and Hervé Bredin},
  title={{Powerset multi-class cross entropy loss for neural speaker diarization}},
  year=2023,
  booktitle={Proc. INTERSPEECH 2023},
}

@inproceedings{karaev2024cotracker,
  title={Cotracker: It is better to track together},
  author={Karaev, Nikita and Rocco, Ignacio and Graham, Benjamin and Neverova, Natalia and Vedaldi, Andrea and Rupprecht, Christian},
  booktitle={European Conference on Computer Vision},
  pages={18--35},
  year={2024},
  organization={Springer}
}

@inproceedings{chung2016out,
  title={Out of time: automated lip sync in the wild},
  author={Chung, Joon Son and Zisserman, Andrew},
  booktitle={Asian Conference on Computer Vision},
  pages={251--263},
  year={2016},
  organization={Springer}
}

@inproceedings{deng2019arcface,
  title={Arcface: Additive angular margin loss for deep face recognition},
  author={Deng, Jiankang and Guo, Jia and Xue, Niannan and Zafeiriou, Stefanos},
  booktitle={Proceedings of the IEEE/CVF Conference on Computer Vision and Pattern Recognition},
  pages={4690--4699},
  year={2019}
}

@article{loshchilov2017decoupled,
  title={Decoupled weight decay regularization},
  author={Loshchilov, Ilya and Hutter, Frank},
  journal={arXiv preprint arXiv:1711.05101},
  year={2017}
}

@article{heusel2017gans,
  title={Gans trained by a two time-scale update rule converge to a local nash equilibrium},
  author={Heusel, Martin and Ramsauer, Hubert and Unterthiner, Thomas and Nessler, Bernhard and Hochreiter, Sepp},
  journal={Advances in Neural Information Processing Systems},
  volume={30},
  year={2017}
}

@article{unterthiner2019fvd,
  title={FVD: A new metric for video generation},
  author={Unterthiner, Thomas and Van Steenkiste, Sjoerd and Kurach, Karol and Marinier, Rapha{\"e}l and Michalski, Marcin and Gelly, Sylvain},
  year={2019}
}

@article{wei2024aniportrait,
  title={Aniportrait: Audio-driven synthesis of photorealistic portrait animation},
  author={Wei, Huawei and Yang, Zejun and Wang, Zhisheng},
  journal={arXiv preprint arXiv:2403.17694},
  year={2024}
}

@article{wang2025fantasytalking,
   title={FantasyTalking: Realistic Talking Portrait Generation via Coherent Motion Synthesis},
   author={Wang, Mengchao and Wang, Qiang and Jiang, Fan and Fan, Yaqi and Zhang, Yunpeng and Qi, Yonggang and Zhao, Kun and Xu, Mu},
  journal={Proceedings of the 33th ACM International Conference on Multimedia},
  year={2025}
}

@misc{jimeng_platform,
  author = {ByteDance},
  title = {Jimeng Platform},
  howpublished = {\url{https://jimeng.jianying.com/ai-tool/home?type=digitalHuman}},
  year = {2025}
}

@article{kolors,
  title={Kolors: Effective Training of Diffusion Model for Photorealistic Text-to-Image Synthesis},
  author={Kolors Team},
  journal={arXiv preprint},
  year={2024}
}

@inproceedings{takahashi2017multi,
  title={Multi-scale multi-band densenets for audio source separation},
  author={Takahashi, Naoya and Mitsufuji, Yuki},
  booktitle={2017 IEEE Workshop on Applications of Signal Processing to Audio and Acoustics (WASPAA)},
  pages={21--25},
  year={2017},
  organization={IEEE}
}

@article{li2024latentsync,
  title={LatentSync: Taming Audio-Conditioned Latent Diffusion Models for Lip Sync with SyncNet Supervision},
  author={Li, Chunyu and Zhang, Chao and Xu, Weikai and Lin, Jingyu and Xie, Jinghui and Feng, Weiguo and Peng, Bingyue and Chen, Cunjian and Xing, Weiwei},
  journal={arXiv preprint arXiv:2412.09262},
  year={2024}
}

@InProceedings{Zhang_2025_CVPR,
    author    = {Zhang, Ruihan and Yu, Borou and Min, Jiajian and Xin, Yetong and Wei, Zheng and Shi, Juncheng Nemo and Huang, Mingzhen and Kong, Xianghao and Xin, Nix Liu and Jiang, Shanshan and Bahuguna, Praagya and Chan, Mark and Hora, Khushi and Yang, Lijian and Liang, Yongqi and Bian, Runhe and Liu, Yunlei and Valencia, Isabela Campillo and Tredinick, Patricia Morales and Kozlov, Ilia and Jiang, Sijia and Huang, Peiwen and Chen, Na and Liu, Xuanxuan and Rao, Anyi},
    title     = {Generative AI for Film Creation: A Survey of Recent Advances},
    booktitle = {Proceedings of the IEEE/CVF Conference on Computer Vision and Pattern Recognition (CVPR) Workshops},
    month     = {June},
    year      = {2025},
    pages     = {6266-6278}
}

@article{kong2025profashion,
  title={ProFashion: Prototype-guided Fashion Video Generation with Multiple Reference Images},
  author={Kong, Xianghao and Qi, Qiaosong and Wang, Yuanbin and Rao, Anyi and Chen, Biaolong and Zhang, Aixi and Liu, Si and Jiang, Hao},
  journal={arXiv preprint arXiv:2505.06537},
  year={2025}
}

@article{zhang2025zero,
  title={Zero-shot 3D-Aware Trajectory-Guided image-to-video generation via Test-Time Training},
  author={Zhang, Ruicheng and Zhou, Jun and Xu, Zunnan and Liu, Zihao and Huang, Jiehui and Zhang, Mingyang and Sun, Yu and Li, Xiu},
  journal={arXiv preprint arXiv:2509.06723},
  year={2025}
}

@article{kong2025taming,
  title={Taming Flow-based I2V Models for Creative Video Editing},
  author={Kong, Xianghao and Chen, Hansheng and Guo, Yuwei and Zhang, Lvmin and Wetzstein, Gordon and Agrawala, Maneesh and Rao, Anyi},
  journal={arXiv preprint arXiv:2509.21917},
  year={2025}
}

@article{hu2022lora,
  title={Lora: Low-rank adaptation of large language models.},
  author={Hu, Edward J and Shen, Yelong and Wallis, Phillip and Allen-Zhu, Zeyuan and Li, Yuanzhi and Wang, Shean and Wang, Lu and Chen, Weizhu and others},
  journal={ICLR},
  volume={1},
  number={2},
  pages={3},
  year={2022}
}

@article{bai2025recammaster,
  title={Recammaster: Camera-controlled generative rendering from a single video},
  author={Bai, Jianhong and Xia, Menghan and Fu, Xiao and Wang, Xintao and Mu, Lianrui and Cao, Jinwen and Liu, Zuozhu and Hu, Haoji and Bai, Xiang and Wan, Pengfei and others},
  journal={arXiv preprint arXiv:2503.11647},
  year={2025}
}

@inproceedings{bahmani2025ac3d,
  title={Ac3d: Analyzing and improving 3d camera control in video diffusion transformers},
  author={Bahmani, Sherwin and Skorokhodov, Ivan and Qian, Guocheng and Siarohin, Aliaksandr and Menapace, Willi and Tagliasacchi, Andrea and Lindell, David B and Tulyakov, Sergey},
  booktitle={Proceedings of the Computer Vision and Pattern Recognition Conference},
  pages={22875--22889},
  year={2025}
}

@inproceedings{zhang2025learning,
  title={Learning Implicit Features with Flow-Infused Transformations for Realistic Virtual Try-On},
  author={Zhang, Delong and Huang, Qiwei and Sun, Yang and Liu, Yuanliu and Zheng, Wei-Shi and Xiong, Pengfei and Zhang, Wei},
  booktitle={Proceedings of the IEEE/CVF International Conference on Computer Vision},
  pages={18736--18745},
  year={2025}
}

@article{ye2025unic,
  title={UNIC: Unified In-Context Video Editing},
  author={Ye, Zixuan and He, Xuanhua and Liu, Quande and Wang, Qiulin and Wang, Xintao and Wan, Pengfei and Zhang, Di and Gai, Kun and Chen, Qifeng and Luo, Wenhan},
  journal={arXiv preprint arXiv:2506.04216},
  year={2025}
}

@inproceedings{lu2025adversarial,
  title={Adversarial distribution matching for diffusion distillation towards efficient image and video synthesis},
  author={Lu, Yanzuo and Ren, Yuxi and Xia, Xin and Lin, Shanchuan and Wang, Xing and Xiao, Xuefeng and Ma, Andy J and Xie, Xiaohua and Lai, Jian-Huang},
  booktitle={Proceedings of the IEEE/CVF International Conference on Computer Vision},
  pages={16818--16829},
  year={2025}
}

@article{ren2024hyper,
  title={Hyper-sd: Trajectory segmented consistency model for efficient image synthesis},
  author={Ren, Yuxi and Xia, Xin and Lu, Yanzuo and Zhang, Jiacheng and Wu, Jie and Xie, Pan and Wang, Xing and Xiao, Xuefeng},
  journal={Advances in Neural Information Processing Systems},
  volume={37},
  pages={117340--117362},
  year={2024}
}

@inproceedings{lu2024coarse,
  title={Coarse-to-fine latent diffusion for pose-guided person image synthesis},
  author={Lu, Yanzuo and Zhang, Manlin and Ma, Andy J and Xie, Xiaohua and Lai, Jianhuang},
  booktitle={Proceedings of the IEEE/CVF Conference on Computer Vision and Pattern Recognition},
  pages={6420--6429},
  year={2024}
}

@inproceedings{ye2025stylemaster,
  title={Stylemaster: Stylize your video with artistic generation and translation},
  author={Ye, Zixuan and Huang, Huijuan and Wang, Xintao and Wan, Pengfei and Zhang, Di and Luo, Wenhan},
  booktitle={Proceedings of the Computer Vision and Pattern Recognition Conference},
  pages={2630--2640},
  year={2025}
}
